%% file: neurips_2025.tex
\documentclass{article}
\usepackage[final]{neurips_2025}
\usepackage{neurips_2025}

\usepackage[utf8]{inputenc} 
\usepackage[T1]{fontenc}    
\usepackage{hyperref}       
\usepackage{url}            
\usepackage{booktabs}       
\usepackage{amsfonts}       
\usepackage{nicefrac}       
\usepackage{microtype}      
\usepackage{xcolor}         
\usepackage{graphicx}
\usepackage{wrapfig}
\usepackage{amsmath}
\usepackage{multirow}
\usepackage{subfig}
\usepackage{enumitem}
\usepackage[figuresright]{rotating}
\usepackage[normalem]{ulem}
\useunder{\uline}{\ul}{}
\PassOptionsToPackage{numbers, square}{natbib}

\definecolor{color1}{HTML}{006EB8}
\hypersetup{
  colorlinks   = true,
  urlcolor     = color1,
  linkcolor    = color1,
  citecolor   = color1
}

\title{\textit{VL-SAE}: Interpreting and Enhancing Vision-Language Alignment with a Unified Concept Set}

\author{
  Shufan Shen$^{1,2}$\quad
  Junshu Sun$^{1,2}$\quad
  Qingming Huang$^{1,2}$\quad
  Shuhui Wang$^{1}$\thanks{Corresponding author.}\and
  $^{1}$Key Lab of Intell. Info. Process., Inst. of Comput. Tech., CAS \\
  $^{2}$University of Chinese Academy of Sciences\and
  \texttt{\{shenshufan22z, sunjunshu21s, wangshuhui\}@ict.ac.cn} \quad
  \texttt{qmhuang@ucas.ac.cn}
}

\begin{document}

\maketitle

\input{Styles/sections/0_abstract}
\input{Styles/sections/1_introduction}
\input{Styles/sections/2_relatedwork}

\input{Styles/sections/3_methodology}
\input{Styles/sections/4_experiments}
\input{Styles/sections/5_conclusion}


\clearpage

\section*{Acknowledgement and Disclosure of Funding}
This work was supported in part by the National Key R$\&$D Program of China under Grant 2023YFC2508704, in part by the National Natural Science Foundation of China 62236008, and in part by the Fundamental Research Funds for the Central Universities.
The authors would like to thank Siqi Zhang, Yue Wu, and the anonymous reviewers for their constructive comments and suggestions that improved this manuscript.

{
\bibliographystyle{plain}
\bibliography{neurips_2025}
}

\newpage

\appendix

\input{Styles/sections/6_appendix}

\newpage
\section*{NeurIPS Paper Checklist}

\begin{enumerate}

\item {\bf Claims}
    \item[] Question: Do the main claims made in the abstract and introduction accurately reflect the paper's contributions and scope?
    \item[] Answer: \answerYes{} 
    \item[] Justification: The main claims made in the abstract and introduction accurately reflect the paper’s contributions and scope.

\item {\bf Limitations}
    \item[] Question: Does the paper discuss the limitations of the work performed by the authors?
    \item[] Answer: \answerYes{} 
    \item[] Justification: We discuss the limitation in Appendix~\ref{subsec: app_limitation}

\item {\bf Theory assumptions and proofs}
    \item[] Question: For each theoretical result, does the paper provide the full set of assumptions and a complete (and correct) proof?
    \item[] Answer: \answerNA{} 
    \item[] Justification: This paper does not include theoretical results.

    \item {\bf Experimental result reproducibility}
    \item[] Question: Does the paper fully disclose all the information needed to reproduce the main experimental results of the paper to the extent that it affects the main claims and/or conclusions of the paper (regardless of whether the code and data are provided or not)?
    \item[] Answer: \answerYes{} 
    \item[] Justification: Please refer to Section~\ref{subsec: details}, Appendix~\ref{sec: app_details} and the codes in the supplementary.

\item {\bf Open access to data and code}
    \item[] Question: Does the paper provide open access to the data and code, with sufficient instructions to faithfully reproduce the main experimental results, as described in supplemental material?
    \item[] Answer: \answerYes{} 
    \item[] Justification: Please refer to the codes in the supplementary.

\item {\bf Experimental setting/details}
    \item[] Question: Does the paper specify all the training and test details (e.g., data splits, hyperparameters, how they were chosen, type of optimizer, etc.) necessary to understand the results?
    \item[] Answer: \answerYes{} 
    \item[] Justification: Please refer to Section~\ref{subsec: details}.

\item {\bf Experiment statistical significance}
    \item[] Question: Does the paper report error bars suitably and correctly defined or other appropriate information about the statistical significance of the experiments?
    \item[] Answer: \answerNo{} 
    \item[] Justification: This paper does not report error bars following the practice of previous studies~\cite{cunningham2023sparse, lim2024sparse}.

\item {\bf Experiments compute resources}
    \item[] Question: For each experiment, does the paper provide sufficient information on the computer resources (type of compute workers, memory, time of execution) needed to reproduce the experiments?
    \item[] Answer: \answerYes{} 
    \item[] Justification: Please refer to Appendix~\ref{sec: app_details}.
    
\item {\bf Code of ethics}
    \item[] Question: Does the research conducted in the paper conform, in every respect, with the NeurIPS Code of Ethics \url{https://neurips.cc/public/EthicsGuidelines}?
    \item[] Answer: \answerYes{} 
    \item[] Justification: This paper conforms with the NeurIPS Code of Ethics.

\item {\bf Broader impacts}
    \item[] Question: Does the paper discuss both potential positive societal impacts and negative societal impacts of the work performed?
    \item[] Answer: \answerNA{} 
    \item[] Justification: There is no societal impact of the work performed. Because it just focuses on the internel mechanism of pre-trained models.
    
\item {\bf Safeguards}
    \item[] Question: Does the paper describe safeguards that have been put in place for responsible release of data or models that have a high risk for misuse (e.g., pretrained language models, image generators, or scraped datasets)?
    \item[] Answer: \answerNA{} 
    \item[] Justification: This paper poses no such risks.

\item {\bf Licenses for existing assets}
    \item[] Question: Are the creators or original owners of assets (e.g., code, data, models), used in the paper, properly credited and are the license and terms of use explicitly mentioned and properly respected?
    \item[] Answer: \answerYes{} 
    \item[] Justification: Please refer to Section~\ref{subsec: details} and Appendix~\ref{sec: app_details}.

\item {\bf New assets}
    \item[] Question: Are new assets introduced in the paper well documented and is the documentation provided alongside the assets?
    \item[] Answer: \answerYes{} 
    \item[] Justification: Please refer to the codes in the supplementary.

\item {\bf Crowdsourcing and research with human subjects}
    \item[] Question: For crowdsourcing experiments and research with human subjects, does the paper include the full text of instructions given to participants and screenshots, if applicable, as well as details about compensation (if any)? 
    \item[] Answer: \answerNA{} 
    \item[] Justification: The paper does not involve crowdsourcing nor research with human subjects.

\item {\bf Institutional review board (IRB) approvals or equivalent for research with human subjects}
    \item[] Question: Does the paper describe potential risks incurred by study participants, whether such risks were disclosed to the subjects, and whether Institutional Review Board (IRB) approvals (or an equivalent approval/review based on the requirements of your country or institution) were obtained?
    \item[] Answer: \answerNA{} 
    \item[] Justification: This paper does not involve crowdsourcing nor research with human subjects.

\item {\bf Declaration of LLM usage}
    \item[] Question: Does the paper describe the usage of LLMs if it is an important, original, or non-standard component of the core methods in this research? Note that if the LLM is used only for writing, editing, or formatting purposes and does not impact the core methodology, scientific rigorousness, or originality of the research, declaration is not required.
    \item[] Answer: \answerNA{} 
    \item[] Justification: LLM is used only for writing, editing, or formatting purposes.

\end{enumerate}

\end{document}

%% file: Styles/sections/0_abstract.tex
\begin{abstract}
The alignment of vision-language representations endows current Vision-Language Models~(VLMs) with strong multi-modal reasoning capabilities. However, the interpretability of the alignment component remains uninvestigated due to the difficulty in mapping the semantics of multi-modal representations into a unified concept set.  
To address this problem, we propose VL-SAE, a sparse autoencoder that encodes vision-language representations into its hidden activations. Each neuron in its hidden layer correlates to a concept represented by semantically similar images and texts, thereby interpreting these representations with a unified concept set. To establish the neuron-concept correlation, we encourage semantically similar representations to exhibit consistent neuron activations during self-supervised training. First, to measure the semantic similarity of multi-modal representations, we perform their alignment in an explicit form based on cosine similarity. 
Second, we construct the VL-SAE with a distance-based encoder and two modality-specific decoders to ensure the activation consistency of semantically similar representations. Experiments across multiple VLMs~(\textit{e.g.}, CLIP, LLaVA) demonstrate the superior capability of VL-SAE in interpreting and enhancing the vision-language alignment. For interpretation, the alignment between vision and language representations can be understood by comparing their semantics with concepts. For enhancement, the alignment can be strengthened by aligning vision-language representations at the concept level, contributing to performance improvements in downstream tasks, including zero-shot image classification and hallucination elimination. Codes are available at \href{https://github.com/ssfgunner/VL-SAE}{https://github.com/ssfgunner/VL-SAE}.
\end{abstract}

%% file: Styles/sections/1_introduction.tex
\section{Introduction}

\begin{figure}[tb]
  \centering
  \includegraphics[width=0.97\linewidth]{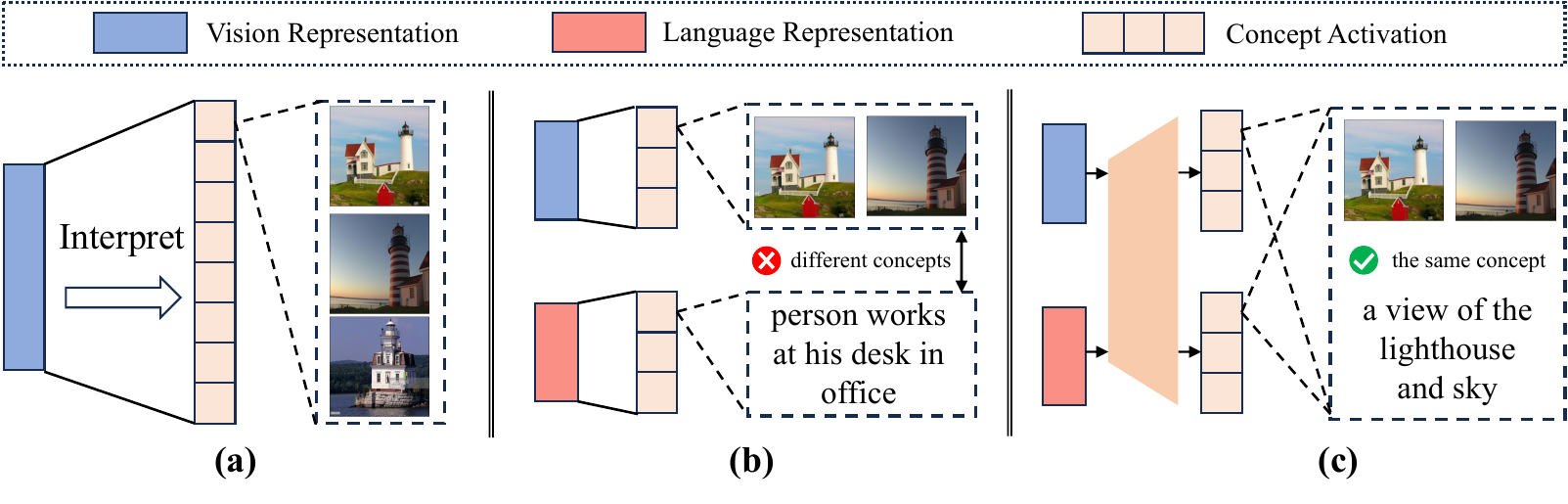}
  \vspace{-0.4cm}
  \caption{(a)~Current interpretation methods are designed for single-modal representations.~(b)~Using current methods for each modality leads to a mismatch in the concept sets, hindering the interpretation of the vision-language alignment.~(c)~We propose the VL-SAE to interpret the alignment mechanism by mapping the representation semantics of both modalities into a unified concept set.}
  \label{fig: intro}
  \vspace{-0.9cm}
\end{figure}

Vision-Language Models~(VLMs)~have demonstrated remarkable capabilities in multi-modal understanding and reasoning, largely attributed to the various training objectives~\cite{jia2021scaling, radford2021learning} and architectures~\cite{tu2023visual, li2023blip, zhang2025cosmo, zhang2025flexvln} that effectively align the semantics of vision and language representations. 
This alignment mechanism serves as a core component of VLMs, enabling them to make predictions by integrating information from both modalities~\cite{radford2021learning, liu2023visual, wang2024qwen2, li2023blip,team2023gemini, grattafiori2024llama}.
Nevertheless, current comprehension of the alignment mechanism remains insufficient~\cite{duan2022multi,shu2025large, parekh2024concept, bhalla2024interpreting}. This limited comprehension hinders our ability to analyze and address the misalignment cases, such as hallucinations~\cite{liu2024survey,leng2024mitigating, huang2024opera}.
To tackle this challenge, existing methods interpret representations of VLMs by mapping their semantics to concepts. However, these methods either solely focus on the vision~\cite{lim2024sparse, bhalla2024interpreting} or the language~\cite{parekh2024concept} representations, as shown in Figure~\ref{fig: intro}\textcolor{color1}{(a)}. 
It remains an open problem to interpret the vision-language alignment, which requires not only understanding the representation semantics but also comparing the semantics of both modalities in an interpretable manner~\cite{shu2025large}.

A straightforward approach to address this problem is to apply existing methods~\cite{shenenhancing, cunningham2023sparse} to map the representation semantics of each modality to concepts separately. 
With these concepts, we can compare the semantics of vision and language representations for interpreting their alignment mechanism.
However, current methods struggle to conduct this comparison with concepts, \textit{as they cannot effectively map the semantics of vision and language representations into a unified concept set}. These methods acquire the concept set in a pre-defined~\cite{kim2018interpretability, shenenhancing, bhalla2024interpreting} or learnable~\cite{lou2025sae, cunningham2023sparse, lim2024sparse} manner.
Pre-defined methods rely on manually constructed concept sets, which often fail to capture the full spectrum of representation semantics~\cite{shenenhancing, yuksekgonul2022post} and suffer from limited scalability due to the need to collect labeled samples for each concept~\cite{kim2018interpretability}.
Learnable methods employ the hidden activations of a pre-trained sparse autoencoder~(SAE) to acquire the concept set, where each hidden neuron correlates to a concept and is defined by the samples that most strongly activate it. 
The neuron-concept correlation is established via end-to-end learning under self-supervision, eliminating the need for concept-specific annotations~\cite{cunningham2023sparse, lim2024sparse}.
Nevertheless, the end-to-end learning introduces uncontrollability into the concept set. 
Due to the uncontrollability, applying two separate SAEs to vision and language representations respectively causes the concept mismatch, \textit{i.e.}, neurons at the same location of different SAEs being correlated with different concepts, as illustrated in Figure~\ref{fig: intro}\textcolor{color1}{(b)}.
As a result, semantically similar vision and language representations can exhibit inconsistent activations under concept mismatch, hindering the comparisons of representation semantics with SAE.

As the concept mismatch stems from applying separate SAEs to different modalities, using a shared SAE for both modalities seems to be a promising solution. By encouraging semantically similar representations to exhibit consistent activations during self-supervised training, the hidden layer neurons can be co-activated by semantically similar images and texts as shown in Figure~\ref{fig: intro}\textcolor{color1}{(c)}, thereby mitigating the concept mismatch. Unfortunately, current SAEs designed for single-modal scenarios fail to achieve this objective for two primary reasons.~\textit{(i)}~In contrast to single-modal representations that use the inner product for measuring semantic similarity, multi-modal representations of different VLMs exhibit varying alignment strategies~\cite{liu2023visual, li2023blip, schuhmann2022laionb}, which complicates the measurement of their semantic similarity.~\textit{(ii)}~Multi-modal representations exhibit modality-specific distributions~\cite{shukor2024implicit}, making it difficult to ensure the activation consistency of semantically similar representations.

In this paper, we propose the VL-SAE, an SAE architecture equipped with an auxiliary autoencoder, to mitigate the concept mismatch. We perform the representation alignment in an explicit form with the auxiliary autoencoder to measure their semantic similarity and then employ VL-SAE to constrain the activation consistency among semantically similar representations. Specifically,~\textit{(i)}~for the alignment strategy, we propose an auxiliary autoencoder that maps the original representations to its intermediate representations. The autoencoder is optimized via contrastive loss~\cite{schuhmann2022laionb} to ensure that vision-language representations with higher semantic similarity exhibit greater cosine similarity between their intermediate representations.~\textit{(ii)}~For the architecture of VL-SAE, we propose an encoder that activates neurons based on the Euclidean distance between normalized representations and neuron weights. This metric satisfies the triangle inequality and is related to the cosine similarity, which correlates with semantic similarity rather than distributional information. These properties ensure that only semantically similar representations can be close to the same weights and activate corresponding neurons. Additionally, we propose separate decoders to capture the distributional information of each modality for representation reconstruction, thereby preventing the encoder from embedding modality-specific information into neuron activations that leads to concept mismatch.

In experiments, we construct VL-SAE on multiple VLMs, including Contrastive VLMs~(CVLMs)~\cite{schuhmann2022laionb} that achieve alignment via retrieval tasks and Large VLMs~(LVLMs)~\cite{liu2023visual, bai2023qwenvl} that rely on question answering tasks. Evaluations of the proposed VL-SAE and current architectures~\cite{cunningham2023sparse, gao2024scaling} demonstrate that VL-SAE possesses superior ability for mapping the semantics of vision and language representations into a unified concept set. 
Subsequently, we utilize VL-SAE to interpret and enhance the vision-language alignment mechanism of VLMs. For interpretation, we explain the model prediction by visualizing the activated concepts of vision and language representations during inference. For enhancement, we propose two strategies to improve the performance of CVLM in zero-shot image classification~\cite{krizhevsky2009learning,wah2011caltech,helber2019eurosat} and LVLM in hallucination elimination~\cite{li2023evaluating} by enhancing the vision-language alignment at the concept level. 
The contributions of our work are as follows:

\begin{itemize}
    \item We propose the VL-SAE, a model to interpret the alignment mechanism of VLMs by mapping the semantics of vision-language representations into a unified concept set.
    \item We apply VL-SAE on multiple widely-used VLMs and demonstrate the superior quality of the concept set learned by VL-SAE in experiments. 
    \item We show that VL-SAE can be employed to interpret the model prediction by visualizing the activated concepts, enhancing the performance of CVLMs on zero-shot image classification, and eliminating the hallucinations of LVLMs. 
\end{itemize}

%% file: Styles/sections/2_relatedwork.tex
\section{Related Work}
\textbf{Vision-Language Models.}
To handle the vision-language reasoning tasks, researchers have developed multiple VLMs~\cite{radford2021learning, liu2023visual, bai2023qwenvl, schuhmann2022laionb, wang2024qwen2, team2023gemini, grattafiori2024llama} by effectively aligning their vision and language representations. According to the pre-training tasks, these models can be categorized into Contrastive VLMs~(CVLMs)~\cite{dong2017learning, schuhmann2022laionb} that utilize retrieval tasks and Large VLMs~(LVLMs)~\cite{liu2023visual,bai2023qwenvl,wang2024qwen2,team2023gemini,grattafiori2024llama} that incorporate Large Language Models~(LLMs)~\cite{touvron2023llama} for pre-training with question answering tasks. For CVLMs, current models include a vision encoder and a text encoder, applied with contrastive training strategies that encourage semantically similar representations to achieve high inner product values~\cite{dong2017learning}. 
For LVLMs, existing methods compose the model with a vision encoder, an LLM, and a light-weight connector that maps the output of the vision encoder to input tokens of the LLM. Vision-language representations are implicitly aligned through pre-training with question answering tasks~\cite{tu2023visual, bai2023qwenvl}.
Despite the various architectures and pre-training paradigms of VLMs, our proposed VL-SAE can interpret and enhance the alignment mechanism of their vision-language representations.

\textbf{Representation Interpretation.} Revealing the semantics of representations stands as one of the primary challenges tackled by interpretable machine learning~\cite{poeta2023concept}. Previous methods~\cite{kim2018interpretability, xu2023energy, espinosa2022concept, shenenhancing} pre-define a concept set and then collect corresponding samples to derive concept vectors in the representation space. The representations are projected onto these vectors for interpretation. These methods are costly due to the requirements for constructing concept sets and collecting relevant samples~\cite{kim2018interpretability}. 
Moreover, these methods encounter challenges in modifying model behavior by adjusting corresponding interpretations, as it is difficult to map these interpretations back to the original representations~\cite{shenenhancing}. Recently, Sparse Autoencoder~(SAE)~\cite{cunningham2023sparse, lim2024sparse, gao2024scaling, li2025sauce} has been recognized as an effective method to interpret representations and modify model behaviors through self-supervised learning. Despite achieving advanced performance, current SAEs remain unsuitable for vision-language representations due to the inconsistency between the learned concept sets of both modalities. 
To address this problem, we propose the VL-SAE with a distance-based encoder and two modality-specific decoders to interpret vision-language representations with a unified concept set.

\textbf{Understanding the Internal Mechanism of VLMs.} There has been an increasing interest in investigating the internal mechanisms of VLMs through the lens of multi-modality~\cite{dang2024explainable}. 
Neuron-based methods~\cite{goh2021multimodal, schwettmann2023multimodal, pan2023finding} reveal the existence of multi-modal neurons that translate vision information to corresponding information in text modality. These neurons are located through important scores like gradient~\cite{schwettmann2023multimodal} and metrics leveraging architecture information~\cite{pan2023finding}. Representation-based methods attempt to interpret the semantics of vision-language representations with concepts. For CVLMs, SpLiCE~\cite{bhalla2024interpreting} and TEXTSPAN~\cite{gandelsman2023interpreting} disentangle the vision representations of CLIP~\cite{radford2021learning} to texts. For LVLMs, Parekh et al.~\cite{parekh2024concept} decompose the token representations to vision-language concepts. SAE-V~\cite{lou2025sae} leverages a sparse autoencoder to token representations for efficient data sampling.
Our method falls within the representation-based category. 
In contrast to existing methods that solely focus on the vision or language representations, our VL-SAE investigates the alignment of vision-language representations by separately mapping their semantics into a unified concept set.

%% file: Styles/sections/3_methodology.tex
\section{Methodology}
We first introduce the architecture of current VLMs~\cite{radford2021learning,liu2023visual, bai2023qwenvl} and SAE~\cite{cunningham2023sparse, lim2024sparse}~(Section~\ref{subsec: preliminaries}). Then, we perform the VLM alignment in an explicit form~(Section~\ref{subsec: alignment}).
Finally, we propose the VL-SAE to map the semantics of vision-language representations into a unified concept set~(Section \ref{subsec: vlsae}).

\subsection{Preliminaries}
\label{subsec: preliminaries}

\noindent\textbf{Contrastive Vision Language Models~(CVLMs)}. Existing CVLMs~\cite{radford2021learning, schuhmann2022laionb} typically consist of a vision encoder and a language encoder. 
Given an input image and text, these encoders independently compute their respective representations. The semantic similarity between the image and text is then estimated through the cosine similarity between their corresponding representations.
By collecting a large number of image-text pairs, CVLMs achieve the vision-language alignment by maximizing the representation similarity between semantically similar image-text pairs.

\noindent\textbf{Large Vision Language Models~(LVLMs)}. A general LVLM architecture~\cite{liu2023visual, li2023blip, bai2023qwenvl} includes a vision encoder, an LLM, and a connector. An input image is first encoded through the vision encoder and then transformed into image tokens via the connector. Subsequently, these image tokens are concatenated with text tokens and provided as input to the LLM for text generation.
With the image-text pairs, the vision-language representations of LVLMs are aligned by pre-training the model to generate textual answers for the text question related to the image content.

To conduct a uniform analysis of both types of models, we utilize the outputs of the vision and language encoders in CVLM for analysis. For LVLM, representations of the image and text tokens in the LLM's hidden layer are averaged across the token axis for analysis.
For convenience, we utilize $\mathbf{x}_v, \mathbf{x}_l \in \mathbb{R}^{d}$ to uniformly represent the vision-language representations of CVLM and LVLM.

\noindent\textbf{Sparse Autoencoder~(SAE)}. SAE is an autoencoder with sparsity constraints on its hidden activations. Given an input $\mathbf{x}\in\mathbb{R}^d$, SAE first transforms it to hidden activations $\mathbf{h}\in\mathbb{R}^h$ using an encoder $E:\mathbb{R}^d\rightarrow\mathbb{R}^h$, and then maps it back to the representation space via a decoder $D:\mathbb{R}^h\rightarrow\mathbb{R}^d$, 
\begin{equation}
    \mathbf{\hat{x}} = D(\mathbf{h}) = D( \sigma(E(\mathbf{x})),
\end{equation}
where $\sigma:\mathbb{R}^h\rightarrow\mathbb{R}^h$ denotes the sparsification function. SAE is trained with the reconstruction loss $\Vert \mathbf{\hat{x}}-\mathbf{x}\Vert^2_2$ in a self-supervised manner. For sparsity constraints, previous methods usually select ReLU as the sparsification function $\sigma$ and integrate the $l_1$ norm of $\mathbf{h}$ into the loss function~\cite{cunningham2023sparse}. Recently, researchers have found that directly adopting the top-k operation as the sparsification function not only eliminates the need for an additional loss term but also demonstrates superior scalability~\cite{gao2024scaling, shenenhancing}.

\subsection{Explicit Representation Alignment}
\label{subsec: alignment}
To ensure the activation consistency among semantically similar vision-language representations for alleviating the concept mismatch, a prerequisite is measuring the semantic similarity of representations from both modalities.
For semantic similarity measurement, more semantically similar vision-language representations have higher cosine similarity in CVLMs~\cite{radford2021learning, schuhmann2022laionb}, which is explicitly constrained via the pre-training objective. 
However, it is difficult to measure the semantic similarity of LVLM representations~\cite{liu2023visual, bai2023qwenvl} due to their implicit alignment mechanism~\cite{shukor2024implicit}. To address this problem, we propose employing an auxiliary autoencoder to convert the implicit alignment mechanism into an explicit alignment mechanism.
Specifically, given an image-text pair, we input them into the VLM to extract representations $(\mathbf{x}_v,\mathbf{x}_l)$. The autoencoder transforms these representations into intermediate representations $\mathbf{x}^e_v, \mathbf{x}^e_l\in\mathbb{R}^{d}$ via encoders $\{E_v, E_l\}$ and then maps them back into the original representations $\mathbf{\hat{x}}_v, \mathbf{\hat{x}}_l\in\mathbb{R}^{d}$ via decoders $\{D_v, D_l\}$,
\begin{equation}
    \mathbf{\hat{x}}_v = D_v(\mathbf{x}^e_v) = D_v(E_v(\mathbf{x}_v)),\ \mathbf{\hat{x}}_l = D_l(\mathbf{x}^e_l) = D_l(E_l(\mathbf{x}_l)). 
\end{equation}
With the contrastive loss InfoNCE~\cite{radford2021learning} and the reconstruction loss, the intermediate representations of image-text pairs can be aligned in cosine similarity while preserving the information of the original representations~(details of InfoNCE are provided in Appendix~\ref{sec: app_details}),
\begin{equation}
    \mathcal{L}(\mathbf{x}_v, \mathbf{x}_l)=\texttt{InfoNCE}(\mathbf{x}^e_v, \mathbf{x}^e_l, \mathbf{x}^{e-}_v, \mathbf{x}^{e-}_l)+\Vert \mathbf{\hat{x}}_v-\mathbf{x}_v\Vert^2_2 + \Vert \mathbf{\hat{x}}_l-\mathbf{x}_l\Vert^2_2,
    \label{eq: alignment}
\end{equation}
where $\mathbf{x}^{e-}_v, \mathbf{x}^{e-}_l$ denote the intermediate representations of negative samples with dissimilar semantics from the given image-text pair. With this autoencoder, we can transform implicitly aligned representations into explicitly aligned intermediate representations, enabling the measurement of semantic similarity between vision and language representations. Note that this component is only utilized for LVLMs with the implicit alignment mechanism. For CVLMs with the explicit alignment mechanism, we directly use their original representations $\mathbf{x}^e_v=\mathbf{x}_v,\mathbf{x}^e_l=\mathbf{x}_l$ for interpretation.
\subsection{Interpreting Vision-Language Representations with a Unified Concept Set}
\label{subsec: vlsae}

\begin{wrapfigure}[21]{r}{0.6\textwidth}
\centering
\includegraphics[width=0.59\textwidth]{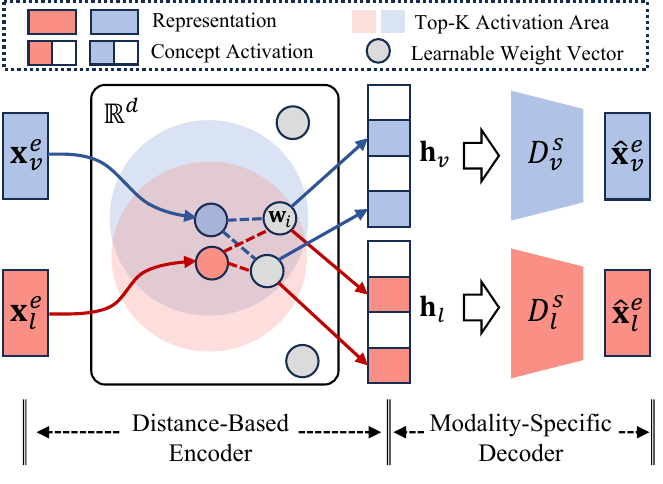}
\vspace{-0.15cm}
\caption{VL-SAE consists of a distance-based encoder that maps semantically similar \textcolor[RGB]{143,170,220}{vision}-\textcolor[RGB]{249,143,133}{language} representations to similar activations, and two modality-specific decoders that reconstruct original representations from these activations.}
\label{fig: method}
\end{wrapfigure}
Based on explicitly aligned representations, we propose the VL-SAE with a distance-based encoder and two modality-specific decoders to ensure the activation consistency of semantically similar representations, thereby alleviating the concept mismatch and interpreting multi-modal representations with a unified concept set.

\noindent\textbf{Encoder}. As the vision-language representations are aligned through cosine similarity, we propose an encoder that activates neurons based on the cosine similarity between the corresponding learnable weight vectors and the input representations. However, since cosine similarity does not satisfy the triangle inequality, two representations with high cosine similarity may not exhibit consistent similarity with the same weight vector, which hinders their activation consistency. To address this problem, we design a distance $g:\mathbb{R}^d\times\mathbb{R}^d\rightarrow[0,2]$ that utilizes the Euclidean distance between normalized inputs. 
Given the representation $\mathbf{x}^e\in\mathbb{R}^d$ and weight vector $\mathbf{w}_i\in\mathbb{R}^d$ corresponding to the $i$-th hidden neuron, their distance is calculated by
\begin{equation}
\label{eq: distance}
    g(\mathbf{x}^e, \mathbf{w}_i) = \left|\left|\frac{\mathbf{x}^e}{\Vert\mathbf{x}^e\Vert_2}-\frac{\mathbf{w}_i}{\Vert\mathbf{w}_i\Vert_2}\right|\right|_2 = \sqrt{2-2cos({\mathbf{x}^e, \mathbf{w}_i})},
\end{equation}
where $cos(\mathbf{x}^e, \mathbf{w}_i)$ denotes the cosine similarity of $\mathbf{x}^e$ and $\mathbf{w}_i$. Since the distance $g$ is a variant of the Euclidean distance, it naturally satisfies the triangle inequality. This indicates that the difference in distance values between the weight vector $\mathbf{w}_i$ and vision-language representations $\mathbf{x}^e_v, \mathbf{x}^e_l$ cannot exceed the distance between the two representations,
\begin{equation}
    \vert g(\mathbf{x}^e_v, \mathbf{w}_i) - g(\mathbf{x}^e_l, \mathbf{w}_i)\vert \le g(\mathbf{x}^e_v, \mathbf{x}^e_l).
\end{equation}
The proposed metric exhibits a negative correlation with cosine similarity, suggesting that as the cosine similarity of vision-language representation increases, the upper bound of the difference in their activations decreases accordingly. We utilize the negative value of $g$ as the activation value, thereby ensuring the positive correlation between the activation value and cosine similarity.
\begin{equation}
    E^{s}(\mathbf{x}^e)[i] = 2-g(\mathbf{x}^e, \mathbf{w}_i) = 2-\sqrt{2-2cos({\mathbf{x}^e, \mathbf{w}_i})},
\end{equation}
where $E^s(\mathbf{x}^e)[i]$ denotes the $i$-th value of $E^s(\mathbf{x}^e)\in\mathbb{R}^h$. We introduce a constant $2$ to ensure the non-negativity of the activation values. For the sparsification function $\sigma$, we employ the top-k function~(\textit{i.e.,} retaining top-k largest values while setting others to zero) considering its strong scalability~\cite{gao2024scaling, shen2024expanding},
\begin{equation}
    \mathbf{h} = \texttt{TopK}(E^s(\mathbf{x}^e)).
\end{equation}
The number of activated neurons $k$ serves as a hyper-parameter. This distance-based encoder bounds the activation discrepancy of semantically similar representations through the triangle inequality, promoting each neuron to be co-activated by semantically similar images and texts.

\noindent\textbf{Decoder.} Despite bounding the activation discrepancy of vision-language representations through the distance-based encoder, it remains insufficient for interpreting representations from distinct distributions. This is because the decoder reconstructs representations solely based on hidden activations. Reconstructing vision-language representations with the same decoder leads to the incorporation of distributional information into the hidden activations during training, thereby reducing the activation consistency of vision-language representations that have similar semantics.

To map the hidden activations into representations of distinct distributions, we propose employing separate decoders for each modality. Given the activations $\mathbf{h}_v, \mathbf{h}_l \in\mathbb{R}^h$ of vision and language representations $\mathbf{x}^e_v, \mathbf{x}^e_l \in\mathbb{R}^d$, two modality-specific decoders $D^s_v, D^s_l$ are utilized to transform these activations back to their original representations, respectively,
\begin{equation}
    \mathbf{\hat{x}}^e_v=D^s_v(\mathbf{h}_v),\ \mathbf{\hat{x}}^e_l=D^s_l(\mathbf{h}_l).
\end{equation}
VL-SAE is trained through the reconstruction loss of visual-language representations,
\begin{equation}
    \mathcal{L}(\mathbf{x}^e_v, \mathbf{x}^e_l)=\Vert \mathbf{\hat{x}}^e_v-\mathbf{x}^e_v\Vert^2_2 + \Vert \mathbf{\hat{x}}^e_l-\mathbf{x}^e_l\Vert^2_2.
\end{equation}
With the separate decoders for storing distributional information, VL-SAE can encode semantically similar vision-language representations to consistent concept activations as interpretations, and then map these activations back to representations of distinct distributions. 

%% file: Styles/sections/4_experiments.tex
\section{Experiments}
In experiments, we first provide the implementation details for constructing VL-SAE based on VLM representations in Section~\ref{subsec: details}. Next, we evaluate the concepts learned by VL-SAE in Section~\ref{subsec: evaluation}. 
Finally, we integrate VL-SAE into the inference process of pre-trained VLMs to interpret and enhance their vision-language alignment mechanisms in Section~\ref{subsec: application}. 
\subsection{Implementation Details for VL-SAE Construction.} 
\label{subsec: details}
\noindent\textbf{VLM Selection.} To demonstrate the broad applicability of the proposed VL-SAE, we build it upon multiple representative VLMs, including CVLMs~(OpenCLIP~\cite{schuhmann2022laionb} with ViT-B/32, ViT-B/16, ViT-L/14, ViT-H/14) pre-trained with retrieval tasks, and LVLMs~(LLaVA1.5~\cite{liu2023visual}, Qwen-VL~\cite{bai2023qwenvl}) pre-trained with question answering tasks. For CVLMs, VL-SAE is constructed using the output from both their vision and language encoders. For LVLMs, VL-SAE is constructed using the output features of the LLM hidden layers~(the $29$-th layer of LLaVA1.5 and the $26$-th layer of Qwen-VL).

\noindent\textbf{Datasets.} We fed the CC3M dataset~\cite{sharma2018conceptual} containing 3 million image-text pairs into the VLM to extract the corresponding vision-language representations. 
These representations are randomly divided into training and test sets at a ratio of 4:1.

\noindent\textbf{Training Strategies.} For LVLMs~\cite{liu2023visual, bai2023qwenvl}, we first train the auxiliary autoencoder for 50 epochs to perform the explicit representation alignment. The training process is configured with a batch size of 2048, a weight decay of 0.01, and a learning rate of 5e-5. Then, the VL-SAE is trained for 10 epochs based on the intermediate representations of the autoencoder with a batch size of 512 and a learning rate of 1e-4. For CVLMs~\cite{schuhmann2022laionb}, we adopt the same strategy as the LVLMs but omit the training of the auxiliary autoencoder because CVLM representations are naturally aligned through cosine similarity.

\subsection{Evaluating the Quality of VL-SAE Concept Set}

\noindent\textbf{Quantitative Evaluation}. 
For the concept set, we evaluate~\textit{(i)}~whether each neuron is activated by semantically similar images and texts, and~\textit{(ii)}~whether diverse semantic concepts are learned across different neurons. These two aspects are quantitatively assessed using the CLIP score \cite{radford2021learning}. Specifically, we first gather the maximally activating images and texts for each neuron of the VL-SAE's hidden layer to describe its semantics. Then, the CLIP score is employed to quantify the semantic similarity between images and texts within the same neuron~(Intra-Similarity) and across different neurons~(Inter-Similarity). The Intra-Similarity evaluates the semantic consistency of images and texts within each concept, while the Inter-Similarity assesses the semantic diversity across the concept set. 
Detailed descriptions of both metrics are presented in Appendix~\ref{subsec: app_construction_details}.

\begin{figure}[tb]
  \centering
  \includegraphics[width=1.0\linewidth]{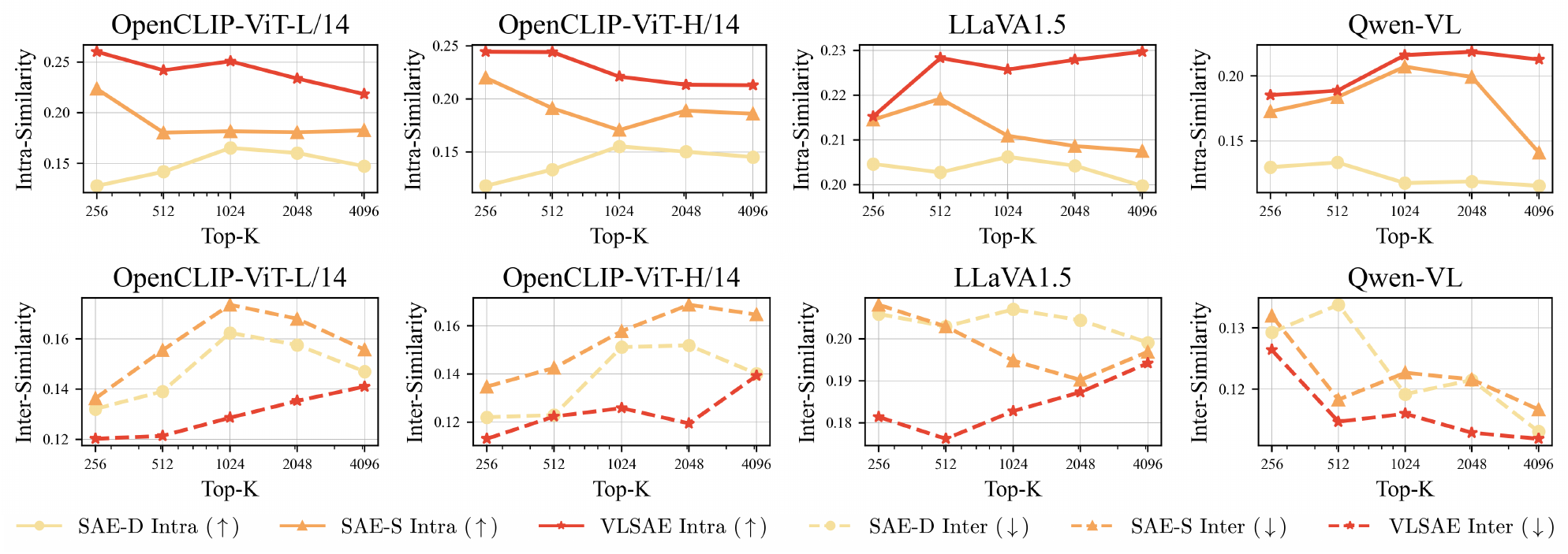}
  \vspace{-0.5cm}
  \caption{Quantitative evaluation of the learned concept set. We compare the VL-SAE with other methods on the consistency between vision and language semantics within the same neuron~(Intra-Similarity) and the semantic diversity across different neurons~(Inter-Similarity) on multiple VLMs.}
  \label{fig: vlsae_eval_quantitive}
  \vspace{-0.3cm}
\end{figure}

For comparison, we consider two baselines using current SAE architectures~\textit{(i)}~SAE-D, which employs distinct SAEs for vision and language representations, and~\textit{(ii)}~SAE-S, which uses a single SAE shared by both modalities. As shown in Figure~\ref{fig: vlsae_eval_quantitive}, VL-SAE exhibits higher intra-similarity and lower inter-similarity compared to these methods, consistently across various VLMs and different numbers of activated concepts. This suggests that \textit{the concepts learned by VL-SAE possess more consistent vision-language semantics and richer semantic diversity than current methods.}

\label{subsec: evaluation}

\begin{figure}[tb]
  \centering
  \includegraphics[width=1.0\linewidth]{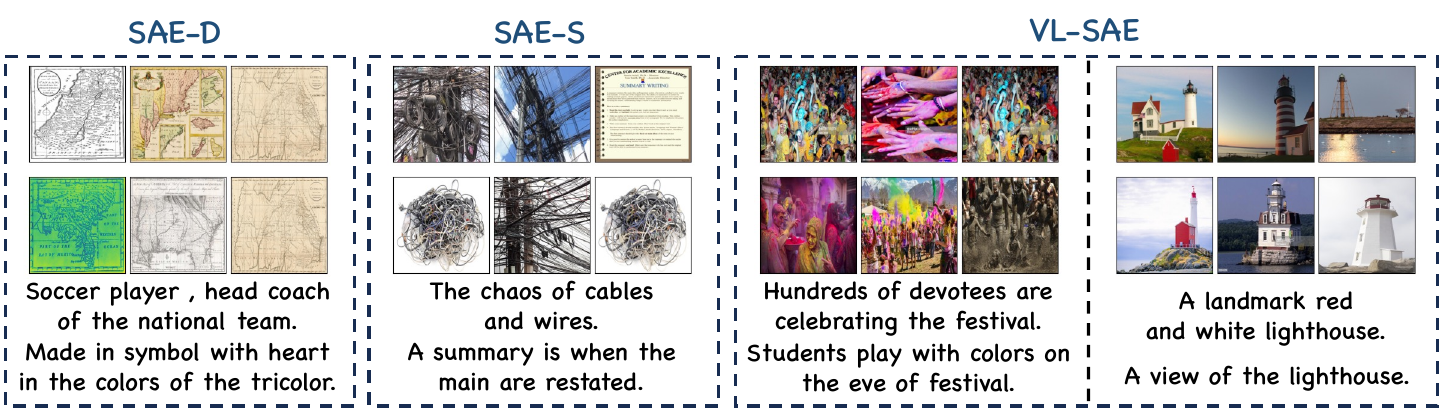}
  \vspace{-0.5cm}
  \caption{Qualitative comparisons among concepts of SAE-D, SAE-S and the proposed VL-SAE. All SAEs are trained with the vision-language representations of LLaVA 1.5.}
  \label{fig: app_comparison}
  \vspace{-0.6cm}
\end{figure}

\noindent\textbf{Qualitative Evaluation.}
We provide qualitative comparisons among SAE-D, SAE-S, and our VL-SAE in Figure~\ref{fig: app_comparison}. SAE-D, which employs separate SAEs for each modality, can extract concepts with consistent semantics within the same modality but encounters the concept mismatch issue as illustrated in Figure~\ref{fig: intro}(b). SAE-S, which utilizes a shared SAE for both modalities, demonstrates a less significant concept mismatch compared to SAE-D. However, semantic inconsistencies still exist, such as the conflation of the semantics of "wire" and "summary tutorials" under a single concept. In contrast, the concepts in VL-SAE consist of samples with consistently aligned semantics.

\subsection{Interpreting and Enhancing the Vision-Language Alignment}
\label{subsec: application}
With the VL-SAE that correlates the semantics of vision and language representations with a unified concept set, we can interpret and enhance the vision-language alignment mechanism in CVLMs and LVLMs by comparing the semantics of both modalities using the concept set. 

\begin{figure}[tb]
  \centering
  \includegraphics[width=0.97\linewidth]{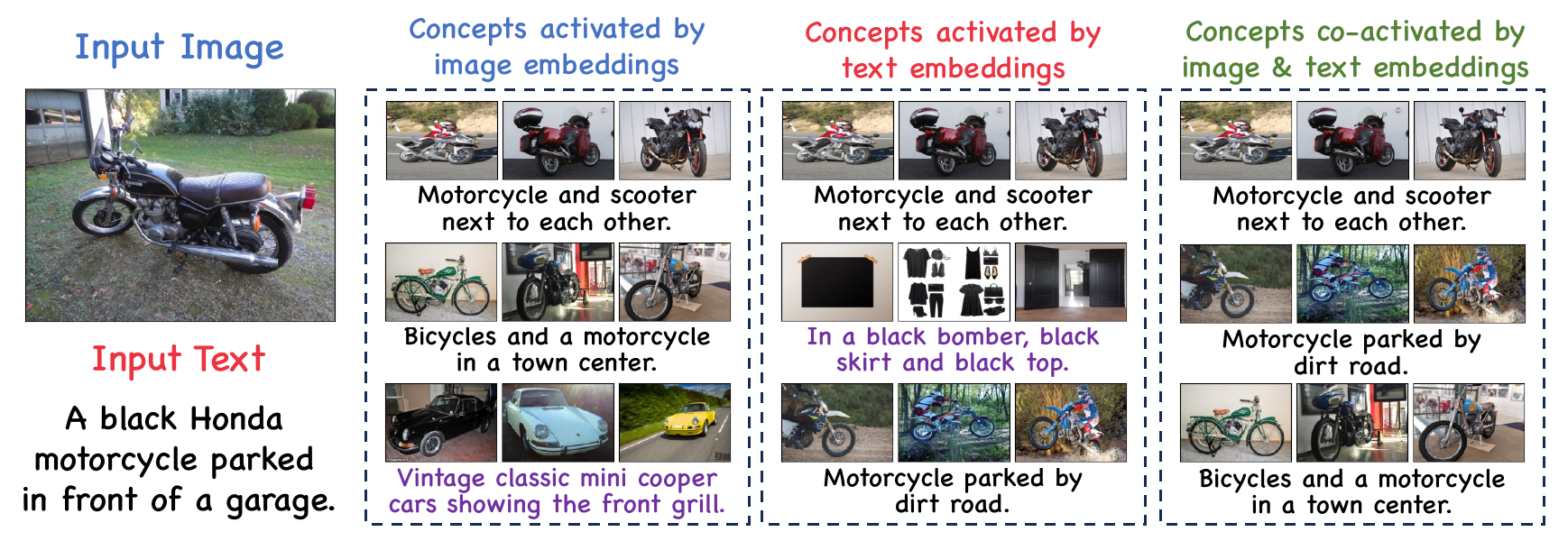}
  \vspace{-0.3cm}
  \caption{Interpreting the vision-language alignment of OpenCLIP-ViT-L/14 with VL-SAE. Given an image-text pair, we visualize the concepts activated by corresponding \textcolor[RGB]{68,114,196}{vision} and \textcolor[RGB]{235,52,65}{language} representations separately, as well as the \textcolor[RGB]{84,130,53}{aligned concepts} co-activated by both modalities. Concepts only related to single-modal representations are highlighted \textcolor[RGB]{112,48,160}{in purple}.}
  \label{fig: vlsae_interpret_retrieval}
  \vspace{-0.3cm}
\end{figure}

\noindent\textbf{Interpreting Vision-Language Alignment of CVLMs}. Given an image-text pair, we separately map the corresponding vision and language representations into the unified concept set through VL-SAE. Figure~\ref{fig: vlsae_interpret_retrieval} depicts several examples by visualizing the concepts activated by representations for each modality and the aligned concepts that are co-activated by both modalities. We observe that~\textit{(i)}~single-modal representations can activate concepts with irrelevant semantics~(\textit{e.g.,} the representations of a motorcycle image activate the concept of a car). This phenomenon reveals the mismatch between representation similarity and semantic similarity in single-modal samples, 
which arises due to contrastive learning's inability to effectively model the relationships among representations within the same modality.
\textit{(ii)}~The aligned concepts exhibit strong semantic similarity to both the given image and text, suggesting that the irrelevant concepts activated by a single-modal representation are not simultaneously activated by representations of the other modality.

\noindent\textbf{Enhancing Vision-Language Alignment of CVLMs.} According to the above interpretations, mapping vision-language representations to a unified concept set enables filtering out the irrelevant concepts activated by single-modal representations. Inspired by this, we enhance the vision-language alignment of CVLMs by aligning their multi-modal representations in concepts. Specifically, for an input image $x_v$ and text $x_l$, their semantic similarity depends not only on the cosine similarities of their representations $\mathbf{x}_v, \mathbf{x}_l$, but also on the cosine similarities of their concept activations $\mathbf{h}_v, \mathbf{h}_l$.
\begin{equation}
\label{eq: classification}
y=cos({\mathbf{x}_{v},\mathbf{x}_{l}})+\alpha_c cos({\mathbf{h}_{v},\mathbf{h}_{l}}),
\end{equation}
where $\alpha_c$ is a task-specific hyperparameter that controls the proportion of concept-based predictions. We enhance the alignment mechanism of multiple CVLMs and conduct evaluations across various zero-shot image classification datasets~\cite{wah2011caltech, krizhevsky2009learning, radford2021learning, cimpoi2014describing, helber2019eurosat, nilsback2008automated, bossard2014food, houben2013detection, deng2012mnist, parkhi2012cats, socher2013recursive, coates2011analysis, xiao2010sun}. As Table~\ref{tab: cvlm_enhance} shows, mapping multi-modal representations to a unified concept set for predictions achieves consistent performance improvement across different models and datasets.

\input{Styles/tables/tab_cvlm_classification}

\begin{figure}[tb]
  \centering
  \includegraphics[width=0.97\linewidth]{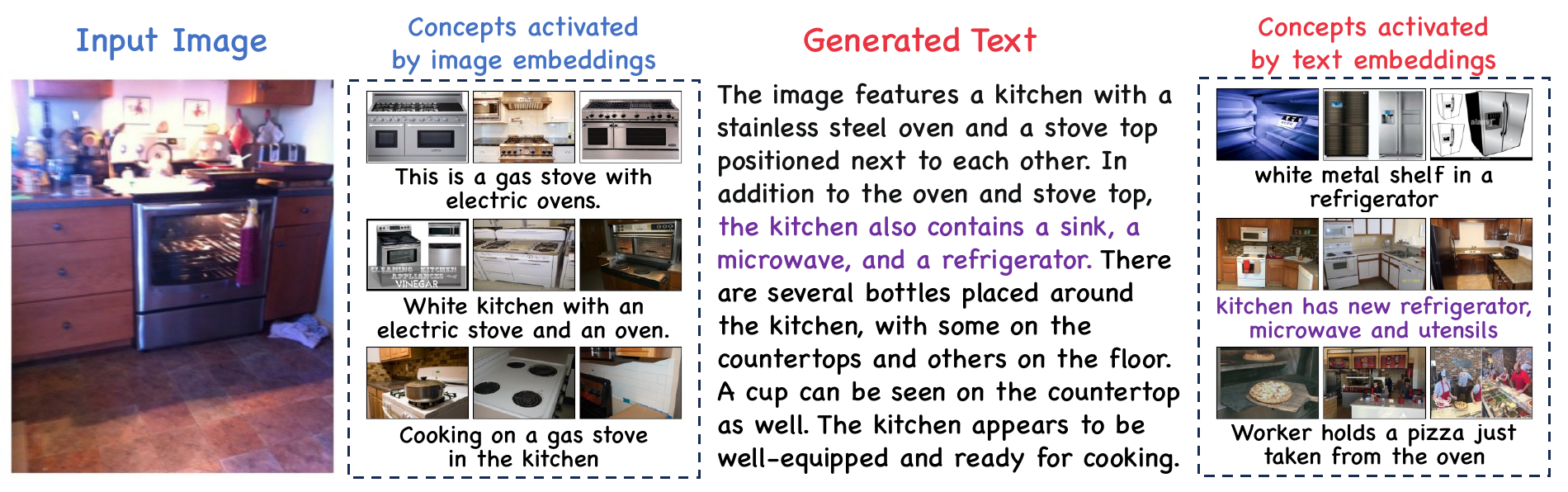}
  \vspace{-0.2cm}
  \caption{Interpreting the vision-language alignment of LLaVA1.5 with VL-SAE. Given an input image and the text generated by LVLMs, we visualize the concepts activated by corresponding \textcolor[RGB]{68,114,196}{vision} and \textcolor[RGB]{235,52,65}{language} representations. The \textcolor[RGB]{112,48,160}{hallucination concept} activated by language representations facilitates a deeper understanding of \textcolor[RGB]{112,48,160}{object hallucinations in the text}.}
  \label{fig: vlsae_interpret_generate}
  \vspace{-0.4cm}
\end{figure}

\noindent\textbf{Interpreting Vision-Language Alignment of LVLMs.}
For LVLMs, we interpret their alignment mechanism by visualizing the concepts activated by the vision and language representations during the text generation process.
As Figure~\ref{fig: vlsae_interpret_generate} shows, given the input image of a kitchen including ovens and stoves, the generated text exhibits object hallucinations of the microwave and refrigerator. By analyzing the activated concepts, we observe that the semantics of both vision and language representations are correlated with the kitchen scene but emphasize different objects. Specifically, the vision representation focuses on ovens and stoves, whereas the language representation is associated with microwaves and refrigerators, which correspond to the hallucinated objects. This phenomenon indicates that the object information in the vision representation is inadequately transferred into the language representation, resulting in hallucinations in the generated text.

\noindent\textbf{Enhancing Vision-Language Alignment of LVLMs}. As the hallucinations of LVLMs are correlated to the misalignment between vision and language representations, we leverage VL-SAE to mitigate hallucinations by enhancing the vision-language alignment at the concept level. Specifically, with the vision and language representations $\mathbf{x}_v, \mathbf{x}_l$, we obtain  a refined language representation $\mathbf{\hat{x}}_l$ by aligning its concept activation $\mathbf{h}_l$ with vision concept activation $\mathbf{h}_v$ through VL-SAE,
\begin{equation}
\label{eq: hallucination}
    \mathbf{\hat{x}}_l = (1-\alpha_l) \mathbf{x}_l+\alpha_l D_l(D^s_l(\mathbf{h}_l+\beta\mathbf{h}_v)).
\end{equation}
The modified language representation $\mathbf{\hat{x}}_l$ is utilized to mitigate hallucinations through contrastive decoding~\cite{leng2024mitigating}. More implementation details are provided in Appendix~\ref{sec: app_details}. Table~\ref{tab: lvlm_enhance} shows the performance comparisons on the POPE~\cite{li2023evaluating} benchmark. As a method for representation interpretation, VL-SAE surpasses the VCD~\cite{leng2024mitigating} that specializes in hallucination elimination. The effectiveness of VL-SAE further highlights the potential of representation interpretation methods to improve model performance on downstream tasks by enhancing the vision-language alignment.

\input{Styles/tables/tab_lvlm_hallucination}

\subsection{Ablation Studies \& Visualizations}
\noindent\textbf{Effects of the VL-SAE Architecture}. 
The ablation studies of the proposed components are provided in Table~\ref{tab: ablation}. 
First, replacing the standard encoder with the cosine-based encoder~(+Cosine-based Encoder), which activates hidden neurons based on cosine similarity, can improve the quality of learned concepts, as vision-language representations are explicitly aligned through cosine similarity. 

\input{Styles/tables/app_abl}
Moreover, replacing the cosine similarity with the distance proposed in Equation~\ref{eq: distance} can improve the concept quality (+Distance-based Encoder). This is because the distance satisfies the triangle inequality, enabling each hidden neuron to be activated by more semantically similar visual-language representations.
Additionally, the improvement achieved by adopting modality-specific decoders stems from mitigating the adverse influence of distributional discrepancies in multi-modal representations on the encoder.
Furthermore, directly training the VL-SAE with original representations of LVLMs~(-Auxiliary Autoencoder) causes the failure of other components, highlighting the necessity of transforming the alignment mechanism to an explicit form.

\input{Styles/tables/abl_data_sparse}

\noindent\textbf{Effects of the Dataset Volume}. 
In Table~\ref{tab: app_data_volume}, we train VL-SAE models with different proportions of the CC3M dataset and report their Intra-Similarity and Inter-Similarity. As the volume of data increases, the Intra-Similarity (0.2029$\rightarrow$0.2299) and Inter-Similarity (0.1597$\rightarrow$0.1220) of the concepts learned by VL-SAE improve accordingly. This suggests that VL-SAE can learn higher-quality concepts by leveraging larger amounts of data in a self-supervised manner.

\noindent\textbf{Effects of the Sparsification Method}.
In Table~\ref{tab: app_topk}, we compare the performance of utilizing Top-k~\cite{shen2024expanding, shenenhancing} and L1 loss~\cite{cunningham2023sparse} for sparsification. The coefficient of L1 loss is set to 1e-4. We find that employing Top-K sparsification yields superior concept quality compared to L1 loss, as the latter lacks precise control over the number of activated hidden neurons. This limitation hinders its ability to effectively balance sparsity in neuron activation with the self-supervised representation reconstruction loss. Our results are consistent with previous studies~\cite{gao2024scaling}, which also indicate that Top-K sparsification is a more effective and scalable approach.

\begin{figure}[tb]
  \centering
  \includegraphics[width=1.0\linewidth]{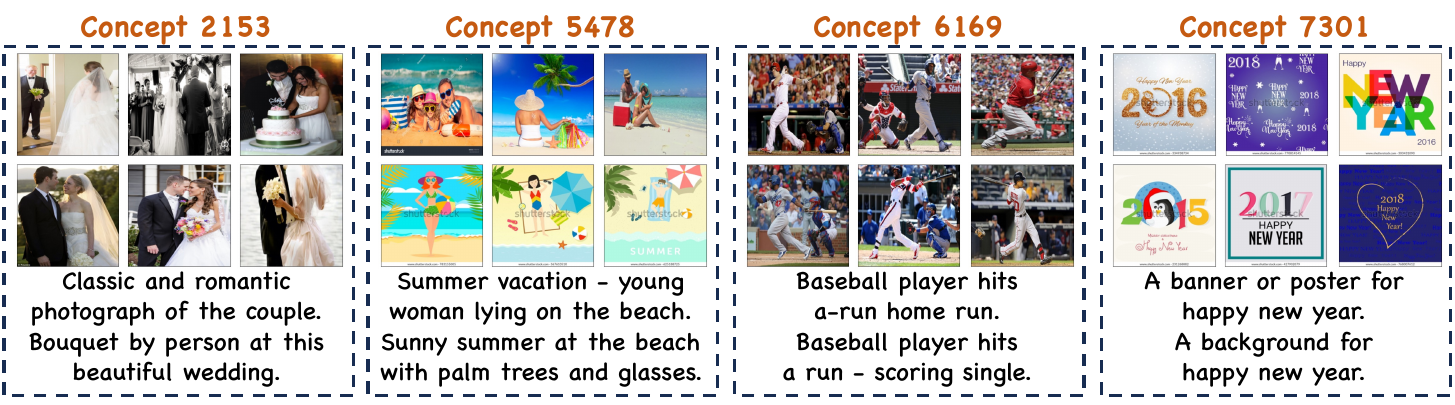}
  \vspace{-0.2cm}
  \caption{Qualitative evaluation of the learned concept set. We provide the maximally activating images and texts for several concepts of VL-SAE trained with the representations of LLaVA 1.5.}
  \label{fig: vlsae_eval_qualitative}
  \vspace{-0.6cm}
\end{figure}

\noindent\textbf{Visualizations of Concepts Learned by VL-SAE}. 
In Figure~\ref{fig: vlsae_eval_qualitative}, we visualize the learned concepts of VL-SAE, which are represented by their maximally activating images and texts. The images and texts of the same concept exhibit highly similar semantics, demonstrating that VL-SAE can effectively alleviate the concept mismatch. Additionally, VL-SAE obtains concepts with abstract semantics rather than focusing exclusively on semantics related to visual appearance. For example, concept 5478 is style-agnostic and activated by both natural and cartoon images, which depict humans lying on a beach. Concept 7301 is color-agnostic and activated by images of happy new year posters. 

%% file: Styles/tables/tab_cvlm_classification.tex
\begin{table}[]
\centering
\footnotesize
\setlength\tabcolsep{4pt}
\caption{Performance of OpenCLIP with varying model sizes in zero-shot image classification tasks.}
\scalebox{1.0}{\begin{tabular}{c|cccccccccccccc|c}
\toprule
& \rotatebox{90}{Caltech101} & \rotatebox{90}{Cifar10} & \rotatebox{90}{Cifar100} & \rotatebox{90}{Country211} & \rotatebox{90}{DTD}  & \rotatebox{90}{Eurosat} & \rotatebox{90}{Flowers} & \rotatebox{90}{Food} & \rotatebox{90}{GTSRB} & \rotatebox{90}{MNIST} & \rotatebox{90}{Pets} & \rotatebox{90}{SST2} & \rotatebox{90}{STL10} & \rotatebox{90}{Sun397} & \rotatebox{90}{Mean Acc.}     \\ \midrule
ViT-B/32       & 86.5       & 93.6    & 75.5     & 16.7       & 55.9 & 47.2    & 71.6    & 82.7 & 49.3  & 69.8  & 90.7 & 57.5 & 96.6  & 68.7   & 68.7          \\
+VL-SAE & 86.7       & 93.8    & 75.7     & 16.7       & 56.9 & 51.0    & 71.8    & 82.8 & 50.3  & 71.8  & 90.7 & 57.9 & 96.9  & 69.4   & \textbf{69.5} \\ \midrule
ViT-B/16       & 86.7       & 94.9    & 76.9     & 20.3       & 56.5 & 52.6    & 71.4    & 86.6 & 46.1  & 66.0  & 90.3 & 59.7 & 97.9  & 70.8   & 69.8          \\
+VL-SAE & 87.2       & 95.1    & 77.0     & 20.4       & 56.7 & 53.2    & 71.6    & 86.8 & 49.1  & 67.8  & 90.9 & 60.4 & 97.9  & 71.4   & \textbf{70.4} \\ \midrule
ViT-L/14       & 87.6       & 95.8    & 78.3     & 24.4       & 61.5 & 57.8    & 74.4    & 88.8 & 51.6  & 64.5  & 92.9 & 60.3 & 98.5  & 74.0   & 72.2          \\
+VL-SAE & 87.9       & 95.9    & 78.6     & 24.5       & 62.8 & 60.3    & 74.5    & 88.8 & 51.9  & 69.1  & 93.0 & 60.6 & 98.6  & 74.3   & \textbf{72.9} \\ \midrule
ViT-H/14       & 88.2       & 97.5    & 84.7     & 29.8       & 67.9 & 72.7    & 80.1    & 92.7 & 58.3  & 72.9  & 94.5 & 64.1 & 98.5  & 75.2   & 76.9          \\
+VL-SAE & 88.5       & 97.7    & 85.0     & 30.0       & 68.4 & 76.7    & 80.1    & 92.9 & 59.5  & 76.1  & 94.7 & 65.2 & 98.6  & 75.3   & \textbf{77.8} \\ \bottomrule
\end{tabular}}
\label{tab: cvlm_enhance}
\vspace{-0.3cm}
\end{table}

%% file: Styles/tables/tab_lvlm_hallucination.tex
\begin{table}[]
\centering
\footnotesize
\setlength\tabcolsep{3.5pt}
\caption{Experimental results on POPE for hallucination elimination.}
\begin{tabular}{c|c|cccc|cccc}
\toprule
\multirow{2}{*}{Setting}     & \multirow{2}{*}{Decoding} & \multicolumn{4}{c|}{LLaVA1.5}            & \multicolumn{4}{c}{Qwen-VL}              \\ \cmidrule(l){3-10} 
                             &                           & Accuracy & Precision & Recall & F1 Score & Accuracy & Precision & Recall & F1 Score \\ \midrule
\multirow{3}{*}{Random}      & Regular                   & 82.93    & 92.01     & 72.13  & 80.87    & 85.20    & 95.99     & 73.47  & 83.23    \\
                             & VCD                       & 85.53    & 93.68     & 76.20  & 84.04    & 86.33    & 96.11     & \textbf{75.73}  & 84.71    \\
                             & VL-SAE                    & \textbf{87.07}    & \textbf{97.28}     & \textbf{76.27}  & \textbf{85.50}    & \textbf{86.47}         & \textbf{97.15}          & 75.13       & \textbf{84.74}         \\ \midrule
\multirow{3}{*}{Popular}     & Regular                   & 81.13    & 87.96     & 72.13  & 79.27    & 83.80    & 94.47     & 71.80  & 81.59    \\
                             & VCD                       & 83.63    & 89.51     & 76.20  & 82.31    & 85.73    & 94.30     & \textbf{76.10}  & 84.21    \\
                             & VL-SAE                    & \textbf{85.87}    & \textbf{94.39}     & \textbf{76.27}  & \textbf{84.37}    & \textbf{85.97}         & \textbf{96.15}          & 74.93       & \textbf{84.22}         \\ \midrule
\multirow{3}{*}{Adversarial} & Regular                   & 78.67    & 83.03     & 72.07  & 77.16    & 82.33    & 89.88     & 72.87  & 80.49    \\
                             & VCD                       & 81.10    & 84.47     & \textbf{76.20}  & 80.13    & 83.77    & 90.03     & \textbf{75.93}  & 82.39    \\
                             & VL-SAE                    & \textbf{83.60}    & \textbf{89.44}     & \textbf{76.20}  & \textbf{82.29}    & \textbf{86.10}         & \textbf{97.29}          & 74.26       & \textbf{84.23}         \\ \bottomrule
\end{tabular}
\label{tab: lvlm_enhance}
\vspace{-0.5cm}
\end{table}

%% file: Styles/tables/app_abl.tex
\begin{wraptable}[10]{r}{9.5cm}
\vspace{-0.4cm}
\caption{Ablation studies of the proposed architecture.}
\vspace{-0.25cm}
\resizebox{1.0\linewidth}{!}{
\setlength{\tabcolsep}{0.5mm}{
\renewcommand\arraystretch{1.0}
\begin{tabular}{@{}l|cc|cc@{}}
\toprule
\multirow{2}{*}{}           & \multicolumn{2}{c|}{OpenCLIP-ViT-H/14}        & \multicolumn{2}{c}{LLaVA1.5} \\ \cmidrule(l){2-5} 
                            & Intra-Sim.~($\uparrow$) & Inter-Sim.~($\downarrow$) & Intra-Sim.~($\uparrow$)  & Inter-Sim.~($\downarrow$)  \\ \midrule
Standard SAE                    & 0.1890           & 0.1688           & 0.2086            & 0.1902            \\ \midrule
+Auxiliary Autoencoder      & -           & -           & 0.2092                 & 0.1908                 \\
+Cosine-based Encoder     & 0.1891           & 0.1518           & 0.2103            & 0.1902            \\
+Distance-based Encoder     & 0.2016           & 0.1357           & 0.2216            & 0.1842            \\
+Modality-specific Decoder & \textbf{0.2134}           & \textbf{0.1149}           & \textbf{0.2257}            & \textbf{0.1828}            \\
-Auxiliary Autoencoder      & -           &  -          & 0.2084                 & 0.2034                 \\ \bottomrule
\end{tabular}}}
\vspace{-0.4cm}
\label{tab: ablation}
\end{wraptable}

%% file: Styles/tables/abl_data_sparse.tex
\begin{table}[t]
\small
	\begin{minipage}[c]{.56\textwidth}
		\centering
\caption{Concept quality of VL-SAE pre-trained with different proportions of the CC3M dataset.}
\resizebox{1.0\linewidth}{!}{
\begin{tabular}{@{}cccccc@{}}
\toprule
Data Percentage      & 10\%   & 30\%   & 50\%   & 70\%   & 100\%  \\ \midrule
Intra-Similarity~($\uparrow$) & 0.2029 & 0.2129 & 0.2196 & 0.2222 & \textbf{0.2299} \\
Inter-Similarity~($\downarrow$) & 0.1597 & 0.1306 & 0.126  & 0.1256 & \textbf{0.1220} \\ \bottomrule
\end{tabular}
}
\label{tab: app_data_volume}
	\end{minipage}
\hfill
	\begin{minipage}[c]{.40\textwidth}%
		\centering
  \caption{Ablation studies of the sparsification method in VL-SAE.}
\resizebox{1.0\linewidth}{!}{
\begin{tabular}{@{}ccc@{}}
\toprule
Method & Intra-Similarity~($\uparrow$) & Inter-Similarity~($\downarrow$) \\ \midrule
L1     & 0.2142           & 0.1809           \\
Top-K  & \textbf{0.2442}           & \textbf{0.1373}           \\ \bottomrule
\end{tabular}
}
\label{tab: app_topk}
	\end{minipage}
\vspace{-0.3cm}
\end{table}

%% file: Styles/sections/5_conclusion.tex
\section{Conclusion}
In this work, we map the semantics of vision-language representations into a unified concept set with VL-SAE, which is trained by encouraging the consistency of its hidden activations among semantically similar representations under self-supervision. To measure the semantic similarity of multi-modal representations, we perform their alignment in an explicit form that correlates the semantic similarity with cosine similarity. To ensure the activation consistency among semantically similar representations, we propose the VL-SAE architecture, including a distance-based encoder and two modality-specific decoders. Experimental results demonstrate the superior ability of VL-SAE to interpret and enhance the alignment mechanism of VLMs. For future work, we will apply the proposed method to more VLMs and investigate its capabilities in additional tasks such as model unlearning and continuous learning. 
For limitations discussion, please refer to Appendix~\ref{subsec: app_limitation}.

%% file: Styles/sections/6_appendix.tex
\renewcommand{\thefigure}{A\arabic{figure}}
\renewcommand{\thetable}{A\arabic{table}}
\renewcommand{\theequation}{A\arabic{equation}}

\begin{figure}[tb]
  \centering
  \includegraphics[width=1.0\linewidth]{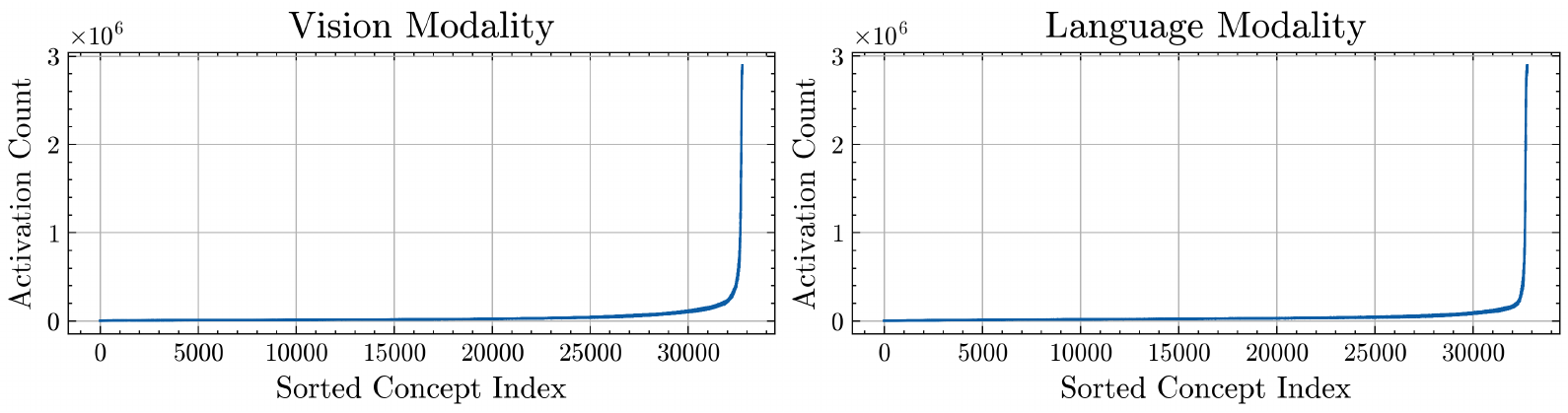}
  \caption{Activation count of each concept in VL-SAE constructed for LLaVA1.5~\cite{liu2023visual}.}
  \label{fig: app_high_concept}
\end{figure}
\input{Styles/tables/app_cvlm_classification_hyper}

\section{More Implementation Details}
\label{sec: app_details}
\subsection{Details of the VL-SAE Construction and Evaluation}
\label{subsec: app_construction_details}

For the InfoNCE loss utilized in Section~\ref{subsec: alignment}, given an image-text pair $(\mathbf{x}_{v}^e, \mathbf{x}_{l}^e)$ and other $N$ with intermediate representations $\{(\mathbf{x}_{v,1}^e, \mathbf{x}_{l,1}^e),...,(\mathbf{x}_{v,N}^e, \mathbf{x}_{l,N}^e)\}$ which denote the $(\mathbf{x}_v^{e-}, \mathbf{x}_l^{e-})$ in Equation~\ref{eq: alignment}, the InfoNCE loss can be represented by 
\begin{equation}
    \texttt{InfoNCE}(\mathbf{x}^e_v, \mathbf{x}^e_l, \mathbf{x}^{e-}_v, \mathbf{x}^{e-}_l) = -\log\left(\frac{\exp(\frac{cos(\theta_{\mathbf{x}^e_v, \mathbf{x}^e_l})}{\tau})}{\sum_{i=1}^N\exp(\frac{cos(\theta_{\mathbf{x}^e_v, \mathbf{x}^e_{l,i}})}{\tau})}\right)-\log\left(\frac{\exp(\frac{cos(\theta_{\mathbf{x}^e_v, \mathbf{x}^e_l})}{\tau})}{\sum_{i=1}^N\exp(\frac{cos(\theta_{\mathbf{x}^e_{v,i}, \mathbf{x}^e_{l}})}{\tau})}\right),
\end{equation}
where $\tau$ denotes the temperature hyperparameter, which is set to 0.07. The InfoNCE loss is computed for image-text pairs within the same batch and aggregated to train the auxiliary autoencoder. 

For the construction of the VL-SAE and auxiliary autoencoder, all the modules except for the distance-based encoder are implemented with a linear layer. We set the hidden ratio to 8 in both CVLMs and LVLMs, indicating that the number of neurons in the hidden layer is 8 times the representation dimension. Unless otherwise specified, the number of activated neurons for each representation is set to 256 for all CVLMs and LVLMs.
For the quantitative evaluation of the concept set, we randomly select several concepts from the autoencoder to compute their intra-similarity and inter-similarity. We calculate these metrics through five random trials, and the average value is adopted to be presented in Figure~\ref{fig: vlsae_eval_quantitive}.
The impact of the number of concepts used for evaluation is shown in Table~\ref{tab: app_concept_num_eval}. The metrics showed no significant differences when more than 100 concepts are used. Since the evaluation needs to be repeated multiple times, we select 100 concepts in each iteration for efficient evaluation.
\input{Styles/tables/app_concept_num_eval}

For the evaluation metrics in Figure~\ref{fig: vlsae_eval_quantitive}, for the $i$-th neuron, we first obtain its maximally activating images and texts. Then, these images and texts are transformed into representations with a pre-trained OpenCLIP-ViT-H/14~\cite{schuhmann2022laionb}. The averaged image and text representations $\mathbf{x}_v^i, \mathbf{x}_l^i$ are utilized to compute the Inter-Similarity and Inter-Similarity metrics. The Inter-Similarity is computed as the average cosine similarity between the image and text representations within the same neuron.
\begin{equation}
    Sim_{intra} = \frac{1}{h} \sum\limits_{i=1}^h cos(\theta_{\mathbf{x}_v^i,\mathbf{x}_l^i}).
\end{equation}
For the Inter-Similarity, it is computed as the average cosine similarity between the image and text representations within the different neurons.
\begin{equation}
    Sim_{inter} = \frac{1}{h(h-1)} \sum\limits_{i=1}^h \sum\limits_{j\neq i}^hcos(\theta_{\mathbf{x}_v^i,\mathbf{x}_l^j}).
\end{equation}
For the resource requirements, the VL-SAE models for different VLMs~\cite{schuhmann2022laionb, liu2023visual, bai2023qwenvl} are all trained using a single NVIDIA GeForce RTX 4090 GPU. For CVLM~\cite{schuhmann2022laionb}, full-precision training is employed to ensure optimal performance and stability. For LVLM~\cite{liu2023visual, bai2023qwenvl}, half-precision~(FP16) training is utilized to effectively reduce resource requirements while maintaining model efficiency.

\subsection{Number of Concepts in VL-SAE}
In Table~\ref{tab: app_concept_num}, we provide the number of concepts in different SAE architectures based on LLaVA 1.5 with 4096 feature dimensions. The number of concepts is upper bounded by the number of hidden neurons in SAEs. Moreover, the number of concepts is influenced by the training process and SAE architectures. This stems from the inevitable existence of dead neurons in SAEs~\cite{gao2024scaling}, which are never activated. The results show that VL-SAE learns more concepts compared to existing methods. 
\input{Styles/tables/app_concept_num}

\subsection{Details of Interpreting Representations with VL-SAE.}
We observe that a subset of concepts in VL-SAE are activated at high frequencies, as Figure~\ref{fig: app_high_concept} shown, \textit{i.e.}, almost all samples activate these concepts. This frequent activation makes it challenging for these concepts to be associated with specific semantics, as shown in Tale~\ref{tab: app_reweighting}. Consequently, in the interpretation process, we divide the target representation's neuron activation by the average activation value of each neuron, thereby mitigating the impact of these high-frequency concepts on the interpretation.
\input{Styles/tables/app_reweighting}

\subsection{Details of the Zero-shot Image Classification Task}
Considering the varying degrees of alignment between the concept set of VL-SAE and the target categories of the downstream task, we experiment with multiple values of $\alpha\in\{0.1, 0.2, 0.3,0.4,0.5,0.6,0.7,0.8,0.9\}$ and select the one that performs best. The selected values are presented in Table~\ref{tab: cvlm_enhance_hyper}.
We observed that models of varying sizes tended to select different $\alpha$ values, whereas for certain datasets like MNIST and Pets, the optimal alpha values remained consistent across different models. Moreover, we observe that the optimal alpha values for some datasets, such as CIFAR-10 and Caltech101, are significantly higher than those for other datasets. This discrepancy stems from the strong correlation between their target categories and the dataset utilized to train the VL-SAE model.

\begin{figure}[tb]
  \centering
  \includegraphics[width=1.0\linewidth]{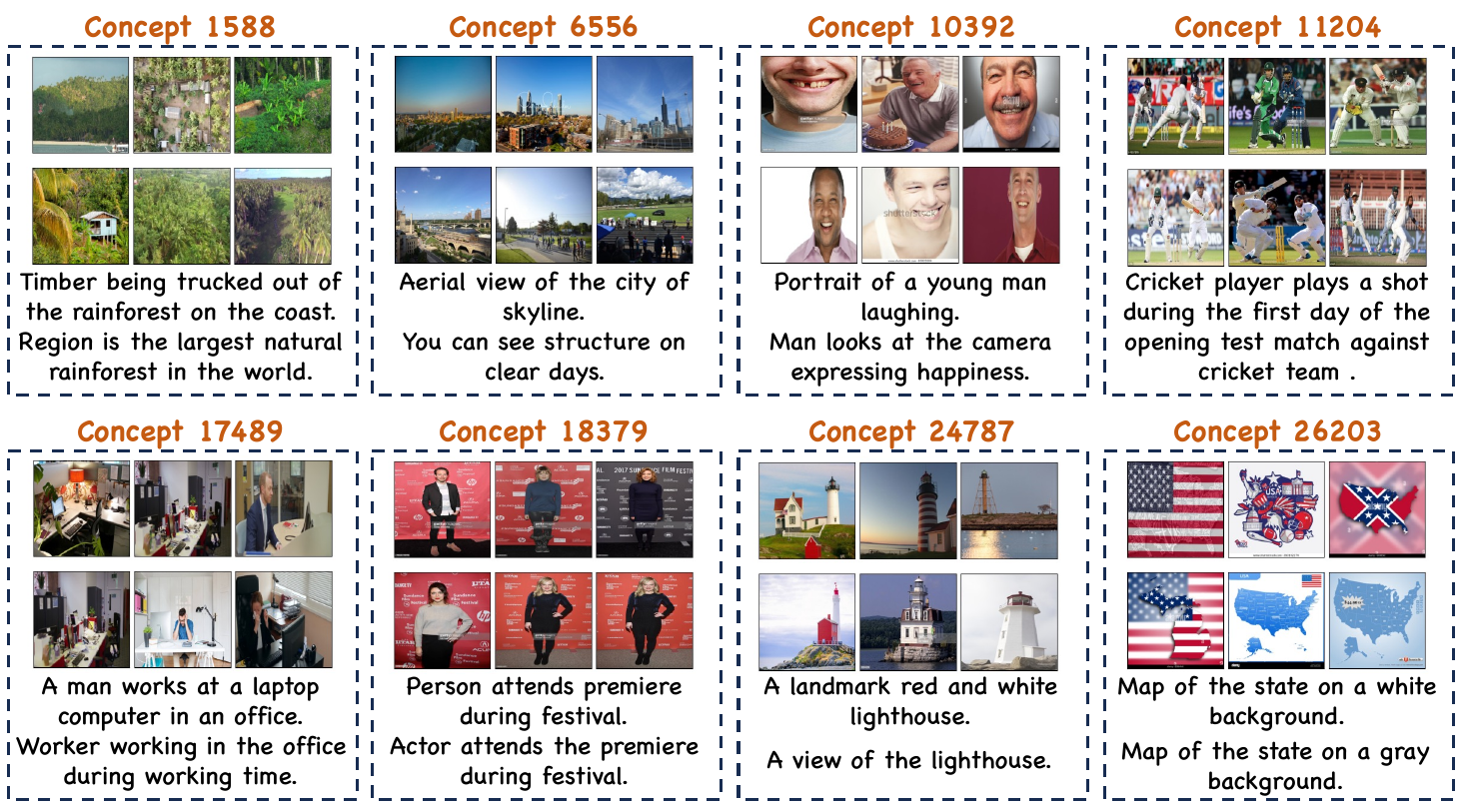}
  \caption{More concepts learned by VL-SAE. We provide the maximally activating images and texts for several concepts of VL-SAE trained for LLaVA1.5.}
  \label{fig: more_concept_app}
\end{figure}

\subsection{Details of the Hallucination Elimination Task}
With the modified representation $\mathbf{\hat{x}}_l$ in Equation~\ref{eq: hallucination}, we adopt the contrastive decoding strategy~\cite{leng2024mitigating} to mitigate the hallucination in LVLMs. Specifically, with the representations of vision and language tokens $\mathbf{x}_{vt}\in\mathbb
R^{N_v\times d},\mathbf{x}_{lt}\in\mathbb
R^{N_l\times d}$, the model generates two distinct output distributions: \textit{(i)} one conditioned on the original representations; and \textit{(ii)} the other conditioned on the original vision representations and the language representations with their mean value across the token dimension replaced by $\mathbf{\hat{x}}_l$.
\begin{equation}
    \mathbf{\hat{x}}_{lt}[i] = \mathbf{x}_{lt}[i] - \mathbf{x}_l + \mathbf{\hat{x}}_l,
\end{equation}
where $\mathbf{x}_{lt}[i]\in\mathbb
R^{d}$ denotes the representation of the $i$-th text token, $\mathbf{x}_l\in\mathbb
R^{d}$ denotes the language representations averaged across the token dimension.
Then, a new contrastive probability distribution is computed by exploiting the differences between the two initially obtained distributions. The new contrastive distribution is formulated as
\begin{equation}
p_{cd}(y \mid \mathbf{x}_{vt}, \mathbf{x}_{lt},\mathbf{\hat{x}}_{lt}) = \operatorname{softmax}[(1-\alpha_{cd}) \operatorname{logit}_\theta(y \mid \mathbf{x}_{vt}, \mathbf{x}_{lt}) +\alpha_{cd} \operatorname{logit}_\theta(y \mid \mathbf{x}_{vt}, \mathbf{\hat{x}}_{lt})].
\end{equation}
Larger $\alpha_{cd}$ values indicate a stronger amplification of differences between the two distributions.
Moreover, we also utilize the adaptive plausibility constraints~\cite{li2022contrastive} that conduct contrastive decoding contingent upon the confidence level $\beta_{cd}$ associated with the output distribution with original representations. In practice, we set $\alpha_{cd}$ and $\beta_{cd}$ to 0.6 and 0.8, respectively. Additionally, the values of $\alpha$ and $\beta$ in Equation~\ref{eq: hallucination} are determined to be 0.7 and 0.9, respectively.

\begin{figure}[tb]
  \centering
  \includegraphics[width=1.0\linewidth]{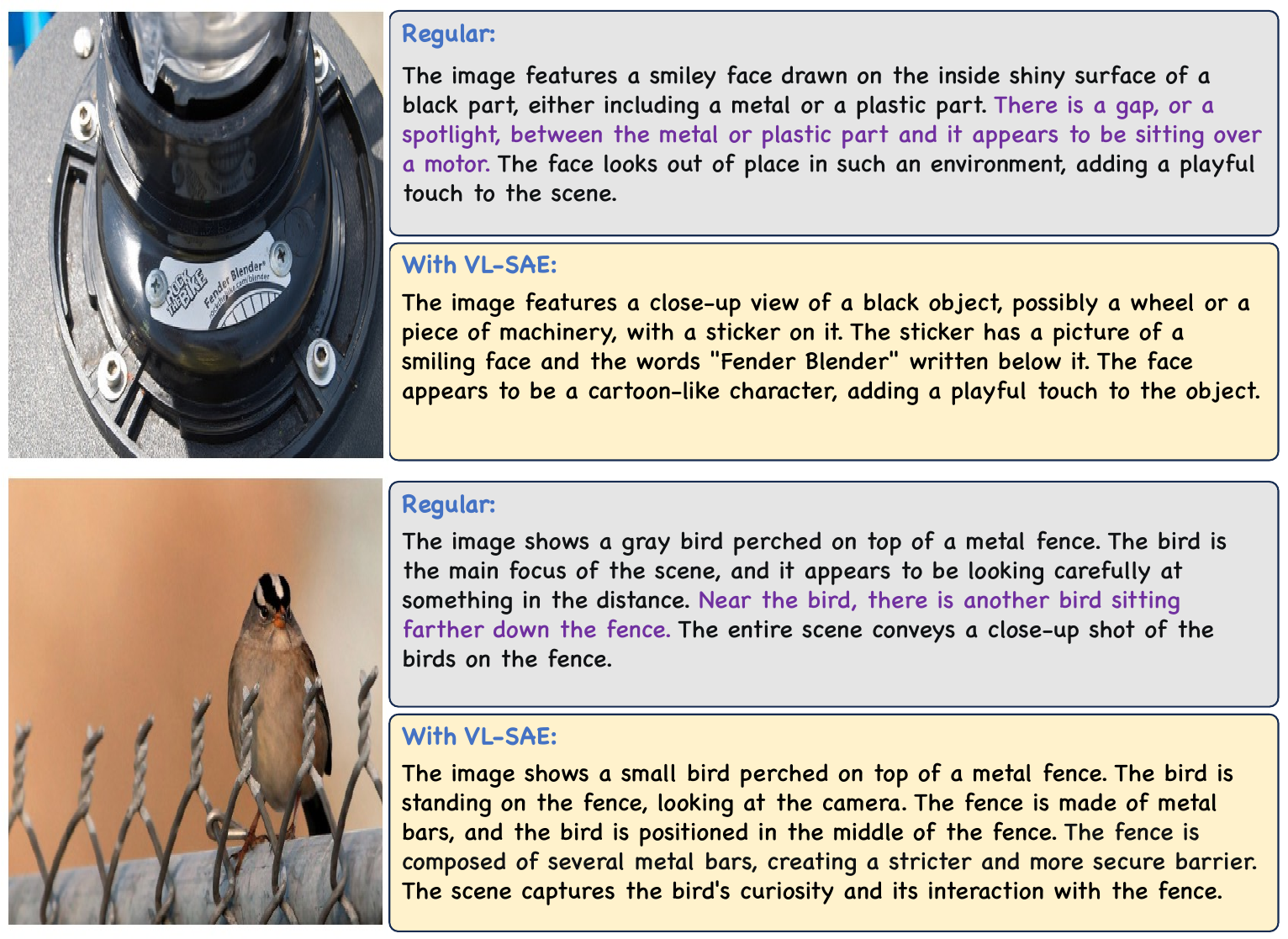}
  \caption{Illustration of generated captions by different decoding methods with LLaVA 1.5 as the backbone model. Hallucinated content is highlighted in \textcolor[RGB]{112,48,160}{purple}.}
  \label{fig: hallucination_visualization_app}
  \vspace{-0.6cm}
\end{figure}

\section{More Experimental Results}

\subsection{Human Evaluation}
We randomly select 100 concepts from SAE-D, SAE-S, and VL-SAE, respectively, and invite 10 participants to evaluate these concepts. For each iteration, participants are required to evaluate the quality of three given concepts and select the one with the highest quality. The method corresponding to the selected concept is awarded one point. The concept quality of these methods is measured by the average score assigned by all participants. As Table~\ref{tab: human} shows, concepts learned by VL-SAE possess higher quality than those learned by other methods. This comparison further highlights the effectiveness of our proposed VL-SAE.
\input{Styles/tables/app_human}

\subsection{More Results for Hallucination Elimination}
Beyond the “Yes-or-No” discriminative evaluations on the POPE datasets, we evaluate the VL-SAE on open-ended caption generation using the CHAIR benchmark~\cite{rohrbach2018object}. We randomly select 500 samples from CHAIR for evaluation and calculate the average of the results obtained from five independent trials. The results in Table~\ref{tab: lvlm_enhance_chair} show the superior performance of VL-SAE over the compared methods. Specifically, VL-SAE reduces object hallucinations in generated captions, as evidenced by lower CHAIR$_S$ and CHAIR$_I$ scores. In addition, VL-SAE enhances the detail of the generated captions, as indicated by higher Recall scores. 
\input{Styles/tables/app_lvlm_enhance_chair}

\subsection{Resource Requirements of VL-SAE}
\noindent\textbf{Training Costs}. Table~\ref{tab: app_training_cost} presents the training cost of VL-SAE. First, as the model size increases, the dimensionality of its learned representations grows, leading to higher computational demands for training VL-SAE. Second, despite this scaling trend, the overall training cost of VL-SAE remains remarkably low. For instance, when compared to LoRA~\cite{hu2021lora}, a widely adopted method known for its computational efficiency, training VL-SAE on OpenCLIP-ViT-B/16 representations incurs significantly fewer FLOPs (0.03G v.s. 1.91G). This minimal computational overhead stems from the small parameter volume in VL-SAE, which is approximately equivalent to that of two linear layers.
\input{Styles/tables/app_training_cost}

\noindent\textbf{Inference Cost}. Table~\ref{tab: app_inference_cost} shows the parameters and computation cost of VL-SAE. For inference speed, we present the influences of VL-SAE in Table~\ref{tab: app_throughput}. The results indicate that VL-SAE has a negligible impact. This is because VL-SAE comprises only two linear layers with significantly smaller parameters than pre-trained VLMs.
\input{Styles/tables/app_inference_cost}
\input{Styles/tables/app_throughput}

\subsection{Additional Ablation Studies}

\noindent\textbf{Selection of Hyperparameter $\alpha_c$.} We first provide accuracy comparisons under three settings: baseline ($\alpha_c=0$), task-agnostic ($\alpha_c=0.1$), and task-specific ($\alpha_c$ determined by tasks). The results are shown in Table~\ref{tab: app_alpha_abl}. 
First, the task-agnostic scheme performs better than the baseline (72.4\% v.s. 72.2\% on ViT-L, and 77.2\% v.s. 76.9\% on ViT-H). This indicates that $\alpha$ does not need to be task-specific for performance improvement and can generalize across tasks.
However, these tasks originate from different domains and exhibit distinct preferences for hyperparameters~\cite{lian2022scaling}, as evidenced by the best performance of the task-specific scheme. If the objective is to achieve optimal performance, the $\alpha$ values of each task should be determined individually,
To determine the $\alpha$ value of each task, we conduct experiments with multiple values $\alpha\in[0,1]$ at intervals of 0.1 and select the one that performs the best. Note that the cost of conducting multiple experiments is very low. For instance, when using the Food-101 dataset, each experiment on a single NVIDIA 4090 GPU takes only approximately 90 seconds.
\input{Styles/tables/app_alpha_abl}

\noindent\textbf{Selection of Hyperparameter $\alpha_l$}. For the hyperparameter $\alpha_l$ in Equation~\ref{eq: hallucination}, We conduct experiments by selecting values at intervals of 0.1 within the range of $[0.5-0.9]$, and the one with the best performance is selected. We report the performance under different values in Table~\ref{tab: app_alphal_abl}. The highest performance across different settings is achieved at $\alpha_l=0.7$.
\input{Styles/tables/app_alphal_abl}

\noindent\textbf{CC3M training}.
In the training process of VL-SAE, the CC3M dataset is employed to integrate VL-SAE as a general plug-in for VLMs, rather than to augment the intrinsic capabilities of these VLMs. In our training strategies, the VLM parameters remain frozen without any adjustments. To further evaluate the impact of CC3M dataset, we provide comparisons between the base model~(CLIP-ViT-B/16), the base model fine-tuned by CC3M, and the base model equipped with VL-SAE in Table~\ref{tab: app_cc3m_abl}. The results show that fine-tuning does not lead to improved performance on downstream tasks. This is because CC3M serves as a subset of the VLM pre-training dataset. It does not provide the base model with any additional task-relevant information.
The superior performance of VL-SAE compared to the fine-tuning model indicates that the improvement stems from the mapping from representations to concepts, rather than the incorporation of CC3M.
\input{Styles/tables/app_cc3m_abl}

\begin{figure}[tb]
  \centering
  \includegraphics[width=1.0\linewidth]{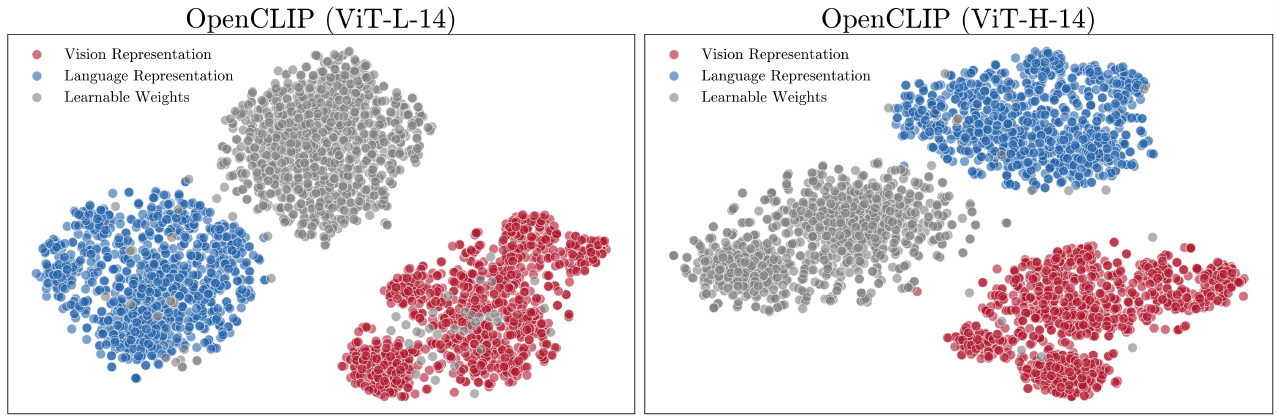}
  \caption{Visualizations of the vision-language representations in the VLMs and the learnable weights in the VL-SAE encoder. Experiments are conducted with t-SNE~\cite{van2008visualizing} for dimension reduction.}
  \label{fig: tsne_app}
\end{figure}

\subsection{Additional Visualizations}
\subsubsection{Concepts Learned by VL-SAE}
We provide more visualization of the concepts learned by VL-SAE in Figure~\ref{fig: more_concept_app}. These results highlight the superior capability of VL-SAE in acquiring a vision-language concept set with rich semantics, such as natural landscapes, urban environments, human expressions, and maps.

\subsubsection{Cases for Hallucination Elimination}
We present qualitative examples of different decoding methods in Figure~\ref{fig: hallucination_visualization_app}. Regular decoding~\cite{liu2023visual} often produces object hallucinations, especially in the latter part of the outputs. In contrast, utilizing VL-SAE effectively mitigates the object hallucinations by enhancing the vision-language alignment at the concept level during the decoding process. 
\subsubsection{Visualization of the Representations and Learnable Weights}
Figure~\ref{fig: tsne_app} illustrates the t-SNE~\cite{van2008visualizing} visualization for the vision-language representations of VLMs and the learnable weights in the encoder of the VL-SAE. We observe that the majority of learnable weights are positioned midway between vision and language representations, rather than being skewed toward either modality. This position enables the corresponding neurons to exhibit consistent activation values across different modalities, which helps map the vision-language representations into a unified concept set.

\begin{figure}[tb]
  \centering
  \includegraphics[width=1.0\linewidth]{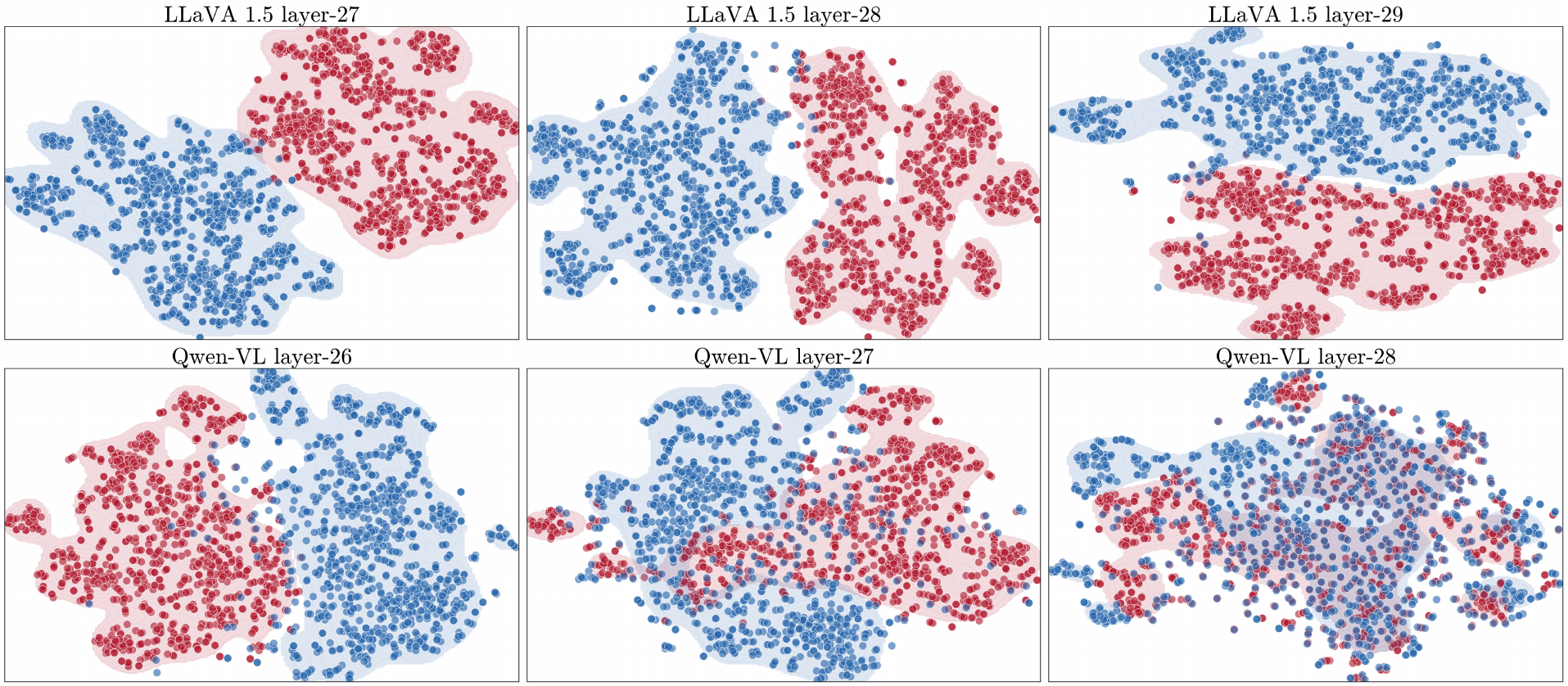}
  \caption{Visualizations of the vision-language representations in different layers of VLMs. Experiments are conducted with t-SNE~\cite{van2008visualizing} for dimension reduction.}
  \label{fig: app_layer_selection}
\end{figure}

\section{Discussion}

\subsection{Criterion of Representation Selection in LVLMs}
We extract the representations from the outputs of the 29-th layer in LLaVA 1.5 and the 26-th layer in Qwen-VL. These layers are chosen because the representations in the top layers tend to be more interpretable in terms of concepts~\cite{jin2024exploring}. Furthermore, at these layers, both vision and language representations preserve their respective information effectively. As shown in Figure~\ref{fig: app_layer_selection}, starting from the 27-th layer of Qwen-VL, the distributions of the two types of representations begin to converge, leading to a potential loss of distinct visual information. This convergence hinders the process of leveraging vision representations to constrain language representations for reducing hallucinations.

\subsection{Metrics for Evaluating the Concept Set}
For the quantitative evaluation of concept collections, we utilized CLIP similarity for automated assessment in Figure~\ref{fig: vlsae_eval_quantitive} and Table~\ref{tab: ablation}. However, the numerical differences in CLIP similarity might not entirely capture the improvements in concept quality achieved by VL-SAE relative to other methods. This is because the numerical gap in CLIP similarity between semantically similar and dissimilar image-text pairs can sometimes be minimal. Therefore, it is important to establish a more comprehensive evaluation mechanism for SAEs based on multi-modal representations in the future.

\subsection{Discussions with attribution-based methods}
We discuss and compare VL-SAE, a concept-based approach, with attribution-based methods~\cite{zhao2024gradient, chefer2021generic} for VLMs, enabling users to select the most appropriate method based on specific application scenarios. 
Attribution-based methods generate a heatmap for each input sample to indicate the importance of individual components of the input data (\textit{e.g.}, image pixels), whereas VL-SAE constructs a concept set across varying samples to interpret the semantics of their hidden representations.

Based on their paradigms, we compare them in the perspective of \textit{intuitiveness}, \textit{expressivity} and \textit{ability to improve model performance}.
\textit{(i)}~Intuitiveness. Compared with VL-SAE, attribution-based methods offer a more intuitive interpretation, as they visualize attribution scores directly on input images.
\textit{(ii)}~Expressivity. Compared with attribution-based methods, VL-SAE exhibits stronger expressivity as it reveals how the model understands the semantics of input data with concepts, rather than simply providing numerical information to identify the important image regions.
\textit{(iii)}~Ability to improve model performance. In contrast to attribution-based methods that only serve as an observer for model prediction, VL-SAE can act as an intervener to improve model performance by modifying hidden representations during inference.

\subsection{Limitation Discussion}
\label{subsec: app_limitation}
Despite successfully correlating the semantics of vision-language representations with a unified concept set, the VL-SAE encounters the same challenges as the SAE architectures for other models~(\textit{e.g.}, LLMs), such as dead neurons~\cite{gao2024scaling}, limited relational modeling, and the impact of high-frequency neurons on interpretation results as shown in Figure~\ref{fig: app_high_concept}. Additionally, the strategy of enhancing the alignment mechanism with VL-SAE is simple and fails to fully unlock the potential of the unified concept set. To address these limitations, we anticipate future work to design SAEs that encourage the neurons to be uniformly activated, incorporate relational modeling~\cite{pmlr-v202-sun23k, NEURIPS2024_93b7e278} among concepts, and enable more effective integration into the prediction processes of VLMs.

\clearpage

%% file: Styles/tables/app_cvlm_classification_hyper.tex
\begin{table}[]
\centering
\footnotesize
\setlength\tabcolsep{6pt}
\caption{The $\alpha$ value of OpenCLIP with varying model sizes in zero-shot image classification tasks.}
\scalebox{1.0}{\begin{tabular}{c|cccccccccccccc}
\toprule
& \rotatebox{90}{Caltech101} & \rotatebox{90}{Cifar10} & \rotatebox{90}{Cifar100} & \rotatebox{90}{Country211} & \rotatebox{90}{DTD}  & \rotatebox{90}{Eurosat} & \rotatebox{90}{Flowers} & \rotatebox{90}{Food} & \rotatebox{90}{GTSRB} & \rotatebox{90}{MNIST} & \rotatebox{90}{Pets} & \rotatebox{90}{SST2} & \rotatebox{90}{STL10} & \rotatebox{90}{Sun397}      \\ \midrule
ViT-B/32 & 0.6        & 0.6     & 0.3      & 0.1        & 0.1 & 0.2     & 0.1     & 0.5  & 0.4   & 1.0   & 0.1  & 0.3  & 0.1   & 0.2    \\
ViT-B/16 & 0.6        & 0.5     & 0.3      & 0.1        & 0.1 & 0.1     & 0.1     & 0.5  & 0.5   & 1.0   & 0.1  & 0.2  & 0.1   & 0.2    \\
ViT-L/14 & 0.6        & 0.4     & 0.3      & 0.3        & 0.1 & 0.3     & 0.2     & 0.1  & 0.2   & 1.0   & 0.1  & 0.1  & 0.9   & 0.1    \\
ViT-H/14 & 0.4        & 0.8     & 0.6      & 0.1        & 0.3 & 0.7     & 0.1     & 0.9  & 0.5   & 1.0   & 0.1  & 0.5  & 0.2   & 0.1    \\ \bottomrule
\end{tabular}}
\label{tab: cvlm_enhance_hyper}
\end{table}

%% file: Styles/tables/app_concept_num_eval.tex
\begin{table}[]
\centering
\caption{Impact of the number of concepts on the evaluation metrics based on OpenCLIP-ViT-B-16.}
\begin{tabular}{@{}cccccc@{}}
\toprule
Concept Num.              & 100             & 200             & 300             & 400             & 500             \\ \midrule
Intra Similarity (SAE-S)  & 0.2200          & 0.2257          & 0.2281          & 0.2245          & 0.2247          \\
Intra Similarity (VL-SAE) & \textbf{0.2445} & \textbf{0.2508} & \textbf{0.2489} & \textbf{0.2476} & \textbf{0.2469} \\ \midrule
Inter Similarity (SAE-S)  & 0.1347          & 0.1293          & 0.1300          & 0.1306          & 0.1308          \\
Inter Similarity (VL-SAE) & \textbf{0.1230} & \textbf{0.1214} & \textbf{0.1182} & \textbf{0.1181} & \textbf{0.1195} \\ \bottomrule
\end{tabular}
\label{tab: app_concept_num_eval}
\end{table}

%% file: Styles/tables/app_concept_num.tex
\begin{table}[]
\centering
\caption{Number of concepts within different SAEs.}
\begin{tabular}{@{}cccc@{}}
\toprule
Method & Hidden Neuron                                             & Dead Neuron & Concept Num. \\ \midrule
SAE-D  & 32768~(4096$\times$8) & 54          & 32714          \\
SAE-S  & 32768~(4096$\times$8) & 46          & 32722          \\
VL-SAE & 32768~(4096$\times$8) & 15          & 32753          \\ \bottomrule
\end{tabular}
\label{tab: app_concept_num}
\end{table}

%% file: Styles/tables/app_reweighting.tex
\begin{table}[]
\centering
\footnotesize
\caption{The concepts of VL-SAE relevant to a kitchen image with and without re-weighting. All concepts are shown in text form. The concept corresponding to the high-frequency neuron is denoted by (*), which exhibits irrelevant semantics with the input kitchen image.}
\begin{tabular}{@{}cc@{}}
\toprule
Before re-weighting                     & After re-weighting                      \\ \midrule
toast popping out of a toaster          & toast popping out of a toaster          \\
hard rock artist performs (*)           & burning on a gas stove in the kitchen   \\
burning on a gas stove in the kitchen   & cans in a refrigerator in a restaurant  \\
cans in a refrigerator in a restaurant  & this is a gas stove with electric ovens \\
this is a gas stove with electric ovens & gas fueled rings on a domestic cooker   \\ \bottomrule
\end{tabular}
\label{tab: app_reweighting}
\end{table}

%% file: Styles/tables/app_human.tex
\begin{table}[]
\caption{Human evaluation of concepts learned by different SAEs.}
\centering
\begin{tabular}{@{}ccc@{}}
\toprule
Method & OpenCLIP-ViT/H & LLaVA 1.5 \\ \midrule
SAE-D  & 0.6            & 0.1       \\
SAE-S  & 33.5           & 35.8      \\
VL-SAE & 65.9           & 64.1      \\ \bottomrule
\end{tabular}
\label{tab: human}
\end{table}

%% file: Styles/tables/app_lvlm_enhance_chair.tex
\begin{table}[]
\centering
\caption{Results on the subset of CHAIR.}
\begin{tabular}{c|ccccc}
\toprule
Model                     & Decoding & CHAIR$_S$~($\downarrow$) & CHAIR$_I$~($\downarrow$) & Recall~($\uparrow$) & Avg. Len \\ \midrule
\multirow{3}{*}{LLaVA1.5} & Regular  & 53.4     & 17.6     & 72.3   & 103.2    \\
                          & VCD      & 55.0     & 16.3     & 76.0   & 102.5    \\
                          & VL-SAE   & \textbf{47.8}     & \textbf{13.3}     & \textbf{76.3}   & 100.7    \\ \midrule
\multirow{3}{*}{Qwen-VL}  & Regular  & 44.6     & 16.1     & 60.7   & 97.2     \\
                          & VCD      & 42.6     & 13.9     & 60.0   & 94.1     \\
                          & VL-SAE   & \textbf{39.6}     & \textbf{10.7}     & \textbf{63.3}   & 94.0     \\ \bottomrule
\end{tabular}
\label{tab: lvlm_enhance_chair}
\end{table}

%% file: Styles/tables/app_training_cost.tex
\begin{table}[]
\centering
\caption{Training cost of VL-SAE based on representations of OpenCLIP.}
\begin{tabular}{@{}cccc@{}}
\toprule
Base Model    & ViT-B/16 & ViT-L/14 & ViT-H/14 \\ \midrule
FLOPs         & 0.03G    & 0.06G    & 0.10G    \\
Training Time & 132s     & 228s     & 446s     \\ \bottomrule
\end{tabular}
\label{tab: app_training_cost}
\end{table}

%% file: Styles/tables/app_inference_cost.tex
\begin{table}[]
\centering
\footnotesize
\caption{Additional parameter and computation cost of VL-SAE.}
\begin{tabular}{@{}ccc|cc@{}}
\toprule
Model          & Total Param. & VL-SAE Param. & Total FLOPs & VL-SAE FLOPs \\ \midrule
LLaVA-1.5      & 7529M        & 467M          & 14.03G      & 0.875G       \\
OpenCLIP-ViT-H & 1011M        & 25M           & 3.15G       & 0.034G       \\ \bottomrule
\end{tabular}
\label{tab: app_inference_cost}
\end{table}

%% file: Styles/tables/app_throughput.tex
\begin{table}[]
\centering
\footnotesize
\caption{Throughput~(sample/s) of the original models and those equipped with VL-SAE.}
\begin{tabular}{ccccc}
\toprule
Method   & OpenCLIP-ViT-B & OpenCLIP-ViT-L & OpenCLIP-ViT-H & LLaVA 1.5 \\ \midrule
Original & 935.32         & 213.62         & 99.45          & 11.34     \\
+VL-SAE  & 935.13         & 213.42         & 99.29          & 10.69  \\  \bottomrule
\end{tabular}
\label{tab: app_throughput}
\end{table}

%% file: Styles/tables/app_alpha_abl.tex
\begin{table}[]
\centering
\caption{Ablation of hyperparameter $\alpha_c$ in Equation~\ref{eq: classification}.}
\begin{tabular}{@{}lcc@{}}
\toprule
Method    & OpenCLIP-ViT-L & OpenCLIP-ViT-H \\ \midrule
baseline~($\alpha_c=0$)      & 72.2           & 76.9           \\
task-agnostic~($\alpha_c=0.1$) & 72.4           & 77.2           \\
task-specific~($\alpha_c$ determined by tasks) & \textbf{72.9}           & \textbf{77.8}           \\ \bottomrule
\end{tabular}
\label{tab: app_alpha_abl}
\end{table}

%% file: Styles/tables/app_alphal_abl.tex
\begin{table}[]
\centering
\caption{Ablation studies on the hyperparameter $\alpha_l$ of Equation~\ref{eq: hallucination}. All experiments are conducted on POPE benchmark with LLaVA 1.5.}
\begin{tabular}{@{}cccccc@{}}
\toprule
Setting     & 0.5   & 0.6   & 0.7   & 0.8   & 0.9   \\ \midrule
Random      & 85.29 & 85.32 & \textbf{85.50} & 85.41 & 85.31 \\
Popular     & 84.09 & 84.15 & \textbf{84.37} & 84.23 & 84.12 \\
Adversarial & 81.92 & 82.10 & \textbf{82.29} & 81.95 & 81.93 \\ \bottomrule
\end{tabular}
\label{tab: app_alphal_abl}
\end{table}

%% file: Styles/tables/app_cc3m_abl.tex
\begin{table}[]
\centering
\caption{Ablation studies on the impact of the CC3M dataset.}
\begin{tabular}{@{}cccc@{}}
\toprule
Method        & Base Model & Base Model+Fine-tuning & Base Model+VL-SAE \\ \midrule
Mean Accuracy & 69.8       & 69.7                   & 70.4              \\ \bottomrule
\end{tabular}
\label{tab: app_cc3m_abl}
\end{table}

%% file: neurips_2025.bbl
\begin{thebibliography}{10}

\bibitem{bai2023qwenvl}
Jinze Bai, Shuai Bai, Shusheng Yang, Shijie Wang, Sinan Tan, Peng Wang, Junyang Lin, Chang Zhou, and Jingren Zhou.
\newblock Qwen-vl: A versatile vision-language model for understanding, localization, text reading, and beyond, 2023.

\bibitem{bhalla2024interpreting}
Usha Bhalla, Alex Oesterling, Suraj Srinivas, Fl{\'{a}}vio~P. Calmon, and Himabindu Lakkaraju.
\newblock Interpreting {CLIP} with sparse linear concept embeddings (splice).
\newblock In Amir Globersons, Lester Mackey, Danielle Belgrave, Angela Fan, Ulrich Paquet, Jakub~M. Tomczak, and Cheng Zhang, editors, {\em Advances in Neural Information Processing Systems 38: Annual Conference on Neural Information Processing Systems 2024, NeurIPS 2024, Vancouver, BC, Canada, December 10 - 15, 2024}, 2024.

\bibitem{bossard2014food}
Lukas Bossard, Matthieu Guillaumin, and Luc Van~Gool.
\newblock Food-101--mining discriminative components with random forests.
\newblock In {\em Computer vision--ECCV 2014: 13th European conference, zurich, Switzerland, September 6-12, 2014, proceedings, part VI 13}, pages 446--461. Springer, 2014.

\bibitem{chefer2021generic}
Hila Chefer, Shir Gur, and Lior Wolf.
\newblock Generic attention-model explainability for interpreting bi-modal and encoder-decoder transformers.
\newblock In {\em Proceedings of the IEEE/CVF international conference on computer vision}, pages 397--406, 2021.

\bibitem{cimpoi2014describing}
Mircea Cimpoi, Subhransu Maji, Iasonas Kokkinos, Sammy Mohamed, and Andrea Vedaldi.
\newblock Describing textures in the wild.
\newblock In {\em 2014 {IEEE} Conference on Computer Vision and Pattern Recognition, {CVPR} 2014, Columbus, OH, USA, June 23-28, 2014}, pages 3606--3613. {IEEE} Computer Society, 2014.

\bibitem{coates2011analysis}
Adam Coates, Andrew Ng, and Honglak Lee.
\newblock An analysis of single-layer networks in unsupervised feature learning.
\newblock In {\em Proceedings of the fourteenth international conference on artificial intelligence and statistics}, pages 215--223. JMLR Workshop and Conference Proceedings, 2011.

\bibitem{cunningham2023sparse}
Hoagy Cunningham, Aidan Ewart, Logan Riggs, Robert Huben, and Lee Sharkey.
\newblock Sparse autoencoders find highly interpretable features in language models.
\newblock {\em ArXiv preprint}, abs/2309.08600, 2023.

\bibitem{dang2024explainable}
Yunkai Dang, Kaichen Huang, Jiahao Huo, Yibo Yan, Sirui Huang, Dongrui Liu, Mengxi Gao, Jie Zhang, Chen Qian, Kun Wang, et~al.
\newblock Explainable and interpretable multimodal large language models: A comprehensive survey.
\newblock {\em ArXiv preprint}, abs/2412.02104, 2024.

\bibitem{deng2012mnist}
Li~Deng.
\newblock The mnist database of handwritten digit images for machine learning research [best of the web].
\newblock {\em IEEE signal processing magazine}, 29(6):141--142, 2012.

\bibitem{dong2017learning}
Xin Dong, Shangyu Chen, and Sinno~Jialin Pan.
\newblock Learning to prune deep neural networks via layer-wise optimal brain surgeon.
\newblock In Isabelle Guyon, Ulrike von Luxburg, Samy Bengio, Hanna~M. Wallach, Rob Fergus, S.~V.~N. Vishwanathan, and Roman Garnett, editors, {\em Advances in Neural Information Processing Systems 30: Annual Conference on Neural Information Processing Systems 2017, December 4-9, 2017, Long Beach, CA, {USA}}, pages 4857--4867, 2017.

\bibitem{duan2022multi}
Jiali Duan, Liqun Chen, Son Tran, Jinyu Yang, Yi~Xu, Belinda Zeng, and Trishul Chilimbi.
\newblock Multi-modal alignment using representation codebook.
\newblock In {\em {IEEE/CVF} Conference on Computer Vision and Pattern Recognition, {CVPR} 2022, New Orleans, LA, USA, June 18-24, 2022}, pages 15630--15639. {IEEE}, 2022.

\bibitem{gandelsman2023interpreting}
Yossi Gandelsman, Alexei~A Efros, and Jacob Steinhardt.
\newblock Interpreting clip's image representation via text-based decomposition.
\newblock {\em ArXiv preprint}, abs/2310.05916, 2023.

\bibitem{gao2024scaling}
Leo Gao, Tom~Dupr{\'e} la~Tour, Henk Tillman, Gabriel Goh, Rajan Troll, Alec Radford, Ilya Sutskever, Jan Leike, and Jeffrey Wu.
\newblock Scaling and evaluating sparse autoencoders.
\newblock {\em ArXiv preprint}, abs/2406.04093, 2024.

\bibitem{goh2021multimodal}
Gabriel Goh, Nick Cammarata, Chelsea Voss, Shan Carter, Michael Petrov, Ludwig Schubert, Alec Radford, and Chris Olah.
\newblock Multimodal neurons in artificial neural networks.
\newblock {\em Distill}, 6(3):e30, 2021.

\bibitem{grattafiori2024llama}
Aaron Grattafiori, Abhimanyu Dubey, Abhinav Jauhri, Abhinav Pandey, Abhishek Kadian, Ahmad Al-Dahle, Aiesha Letman, Akhil Mathur, Alan Schelten, Alex Vaughan, et~al.
\newblock The llama 3 herd of models.
\newblock {\em ArXiv preprint}, abs/2407.21783, 2024.

\bibitem{helber2019eurosat}
Patrick Helber, Benjamin Bischke, Andreas Dengel, and Damian Borth.
\newblock Eurosat: A novel dataset and deep learning benchmark for land use and land cover classification.
\newblock {\em IEEE Journal of Selected Topics in Applied Earth Observations and Remote Sensing}, 12(7):2217--2226, 2019.

\bibitem{houben2013detection}
Sebastian Houben, Johannes Stallkamp, Jan Salmen, Marc Schlipsing, and Christian Igel.
\newblock Detection of traffic signs in real-world images: The german traffic sign detection benchmark.
\newblock In {\em The 2013 international joint conference on neural networks (IJCNN)}, pages 1--8. Ieee, 2013.

\bibitem{hu2021lora}
Edward~J. Hu, Yelong Shen, Phillip Wallis, Zeyuan Allen{-}Zhu, Yuanzhi Li, Shean Wang, Lu~Wang, and Weizhu Chen.
\newblock Lora: Low-rank adaptation of large language models.
\newblock In {\em The Tenth International Conference on Learning Representations, {ICLR} 2022, Virtual Event, April 25-29, 2022}. OpenReview.net, 2022.

\bibitem{huang2024opera}
Qidong Huang, Xiaoyi Dong, Pan Zhang, Bin Wang, Conghui He, Jiaqi Wang, Dahua Lin, Weiming Zhang, and Nenghai Yu.
\newblock Opera: Alleviating hallucination in multi-modal large language models via over-trust penalty and retrospection-allocation.
\newblock In {\em Proceedings of the IEEE/CVF Conference on Computer Vision and Pattern Recognition}, pages 13418--13427, 2024.

\bibitem{jia2021scaling}
Chao Jia, Yinfei Yang, Ye~Xia, Yi{-}Ting Chen, Zarana Parekh, Hieu Pham, Quoc~V. Le, Yun{-}Hsuan Sung, Zhen Li, and Tom Duerig.
\newblock Scaling up visual and vision-language representation learning with noisy text supervision.
\newblock In Marina Meila and Tong Zhang, editors, {\em Proceedings of the 38th International Conference on Machine Learning, {ICML} 2021, 18-24 July 2021, Virtual Event}, volume 139 of {\em Proceedings of Machine Learning Research}, pages 4904--4916. {PMLR}, 2021.

\bibitem{jin2024exploring}
Mingyu Jin, Qinkai Yu, Jingyuan Huang, Qingcheng Zeng, Zhenting Wang, Wenyue Hua, Haiyan Zhao, Kai Mei, Yanda Meng, Kaize Ding, et~al.
\newblock Exploring concept depth: How large language models acquire knowledge and concept at different layers?
\newblock {\em ArXiv preprint}, abs/2404.07066, 2024.

\bibitem{kim2018interpretability}
Been Kim, Martin Wattenberg, Justin Gilmer, Carrie~J. Cai, James Wexler, Fernanda~B. Vi{\'{e}}gas, and Rory Sayres.
\newblock Interpretability beyond feature attribution: Quantitative testing with concept activation vectors {(TCAV)}.
\newblock In Jennifer~G. Dy and Andreas Krause, editors, {\em Proceedings of the 35th International Conference on Machine Learning, {ICML} 2018, Stockholmsm{\"{a}}ssan, Stockholm, Sweden, July 10-15, 2018}, volume~80 of {\em Proceedings of Machine Learning Research}, pages 2673--2682. {PMLR}, 2018.

\bibitem{krizhevsky2009learning}
Alex Krizhevsky, Geoffrey Hinton, et~al.
\newblock Learning multiple layers of features from tiny images.
\newblock 2009.

\bibitem{leng2024mitigating}
Sicong Leng, Hang Zhang, Guanzheng Chen, Xin Li, Shijian Lu, Chunyan Miao, and Lidong Bing.
\newblock Mitigating object hallucinations in large vision-language models through visual contrastive decoding.
\newblock In {\em Proceedings of the IEEE/CVF Conference on Computer Vision and Pattern Recognition}, pages 13872--13882, 2024.

\bibitem{li2023blip}
Junnan Li, Dongxu Li, Silvio Savarese, and Steven C.~H. Hoi.
\newblock {BLIP-2:} bootstrapping language-image pre-training with frozen image encoders and large language models.
\newblock In Andreas Krause, Emma Brunskill, Kyunghyun Cho, Barbara Engelhardt, Sivan Sabato, and Jonathan Scarlett, editors, {\em International Conference on Machine Learning, {ICML} 2023, 23-29 July 2023, Honolulu, Hawaii, {USA}}, volume 202 of {\em Proceedings of Machine Learning Research}, pages 19730--19742. {PMLR}, 2023.

\bibitem{li2025sauce}
Qing Li, Jiahui Geng, Derui Zhu, Fengyu Cai, Chenyang Lyu, and Fakhri Karray.
\newblock Sauce: Selective concept unlearning in vision-language models with sparse autoencoders.
\newblock {\em ArXiv preprint}, abs/2503.14530, 2025.

\bibitem{li2022contrastive}
Xiang~Lisa Li, Ari Holtzman, Daniel Fried, Percy Liang, Jason Eisner, Tatsunori Hashimoto, Luke Zettlemoyer, and Mike Lewis.
\newblock Contrastive decoding: Open-ended text generation as optimization.
\newblock In Anna Rogers, Jordan Boyd-Graber, and Naoaki Okazaki, editors, {\em Proceedings of the 61st Annual Meeting of the Association for Computational Linguistics (Volume 1: Long Papers)}, pages 12286--12312, Toronto, Canada, 2023. Association for Computational Linguistics.

\bibitem{li2023evaluating}
Yifan Li, Yifan Du, Kun Zhou, Jinpeng Wang, Xin Zhao, and Ji-Rong Wen.
\newblock Evaluating object hallucination in large vision-language models.
\newblock In Houda Bouamor, Juan Pino, and Kalika Bali, editors, {\em Proceedings of the 2023 Conference on Empirical Methods in Natural Language Processing}, pages 292--305, Singapore, 2023. Association for Computational Linguistics.

\bibitem{lian2022scaling}
Dongze Lian, Daquan Zhou, Jiashi Feng, and Xinchao Wang.
\newblock Scaling {\&} shifting your features: {A} new baseline for efficient model tuning.
\newblock In Sanmi Koyejo, S.~Mohamed, A.~Agarwal, Danielle Belgrave, K.~Cho, and A.~Oh, editors, {\em Advances in Neural Information Processing Systems 35: Annual Conference on Neural Information Processing Systems 2022, NeurIPS 2022, New Orleans, LA, USA, November 28 - December 9, 2022}, 2022.

\bibitem{lim2024sparse}
Hyesu Lim, Jinho Choi, Jaegul Choo, and Steffen Schneider.
\newblock Sparse autoencoders reveal selective remapping of visual concepts during adaptation.
\newblock {\em ArXiv preprint}, abs/2412.05276, 2024.

\bibitem{liu2024survey}
Hanchao Liu, Wenyuan Xue, Yifei Chen, Dapeng Chen, Xiutian Zhao, Ke~Wang, Liping Hou, Rongjun Li, and Wei Peng.
\newblock A survey on hallucination in large vision-language models.
\newblock {\em ArXiv preprint}, abs/2402.00253, 2024.

\bibitem{liu2023visual}
Haotian Liu, Chunyuan Li, Qingyang Wu, and Yong~Jae Lee.
\newblock Visual instruction tuning.
\newblock In Alice Oh, Tristan Naumann, Amir Globerson, Kate Saenko, Moritz Hardt, and Sergey Levine, editors, {\em Advances in Neural Information Processing Systems 36: Annual Conference on Neural Information Processing Systems 2023, NeurIPS 2023, New Orleans, LA, USA, December 10 - 16, 2023}, 2023.

\bibitem{lou2025sae}
Hantao Lou, Changye Li, Jiaming Ji, and Yaodong Yang.
\newblock Sae-v: Interpreting multimodal models for enhanced alignment.
\newblock {\em ArXiv preprint}, abs/2502.17514, 2025.

\bibitem{nilsback2008automated}
Maria{-}Elena Nilsback and Andrew Zisserman.
\newblock Automated flower classification over a large number of classes.
\newblock In {\em Sixth Indian Conference on Computer Vision, Graphics {\&} Image Processing, {ICVGIP} 2008, Bhubaneswar, India, 16-19 December 2008}, pages 722--729. {IEEE} Computer Society, 2008.

\bibitem{pan2023finding}
Haowen Pan, Yixin Cao, Xiaozhi Wang, Xun Yang, and Meng Wang.
\newblock Finding and editing multi-modal neurons in pre-trained transformers.
\newblock {\em ArXiv preprint}, abs/2311.07470, 2023.

\bibitem{parekh2024concept}
Jayneel Parekh, Pegah Khayatan, Mustafa Shukor, Alasdair Newson, and Matthieu Cord.
\newblock A concept-based explainability framework for large multimodal models.
\newblock In Amir Globersons, Lester Mackey, Danielle Belgrave, Angela Fan, Ulrich Paquet, Jakub~M. Tomczak, and Cheng Zhang, editors, {\em Advances in Neural Information Processing Systems 38: Annual Conference on Neural Information Processing Systems 2024, NeurIPS 2024, Vancouver, BC, Canada, December 10 - 15, 2024}, 2024.

\bibitem{parkhi2012cats}
Omkar~M. Parkhi, Andrea Vedaldi, Andrew Zisserman, and C.~V. Jawahar.
\newblock Cats and dogs.
\newblock In {\em 2012 {IEEE} Conference on Computer Vision and Pattern Recognition, Providence, RI, USA, June 16-21, 2012}, pages 3498--3505. {IEEE} Computer Society, 2012.

\bibitem{poeta2023concept}
Eleonora Poeta, Gabriele Ciravegna, Eliana Pastor, Tania Cerquitelli, and Elena Baralis.
\newblock Concept-based explainable artificial intelligence: A survey.
\newblock {\em ArXiv preprint}, abs/2312.12936, 2023.

\bibitem{radford2021learning}
Alec Radford, Jong~Wook Kim, Chris Hallacy, Aditya Ramesh, Gabriel Goh, Sandhini Agarwal, Girish Sastry, Amanda Askell, Pamela Mishkin, Jack Clark, Gretchen Krueger, and Ilya Sutskever.
\newblock Learning transferable visual models from natural language supervision.
\newblock In Marina Meila and Tong Zhang, editors, {\em Proceedings of the 38th International Conference on Machine Learning, {ICML} 2021, 18-24 July 2021, Virtual Event}, volume 139 of {\em Proceedings of Machine Learning Research}, pages 8748--8763. {PMLR}, 2021.

\bibitem{rohrbach2018object}
Anna Rohrbach, Lisa~Anne Hendricks, Kaylee Burns, Trevor Darrell, and Kate Saenko.
\newblock Object hallucination in image captioning.
\newblock In Ellen Riloff, David Chiang, Julia Hockenmaier, and Jun{'}ichi Tsujii, editors, {\em Proceedings of the 2018 Conference on Empirical Methods in Natural Language Processing}, pages 4035--4045, Brussels, Belgium, 2018. Association for Computational Linguistics.

\bibitem{schuhmann2022laionb}
Christoph Schuhmann, Romain Beaumont, Richard Vencu, Cade Gordon, Ross Wightman, Mehdi Cherti, Theo Coombes, Aarush Katta, Clayton Mullis, Mitchell Wortsman, Patrick Schramowski, Srivatsa Kundurthy, Katherine Crowson, Ludwig Schmidt, Robert Kaczmarczyk, and Jenia Jitsev.
\newblock {LAION-5B:} an open large-scale dataset for training next generation image-text models.
\newblock In Sanmi Koyejo, S.~Mohamed, A.~Agarwal, Danielle Belgrave, K.~Cho, and A.~Oh, editors, {\em Advances in Neural Information Processing Systems 35: Annual Conference on Neural Information Processing Systems 2022, NeurIPS 2022, New Orleans, LA, USA, November 28 - December 9, 2022}, 2022.

\bibitem{schwettmann2023multimodal}
Sarah Schwettmann, Neil Chowdhury, Samuel Klein, David Bau, and Antonio Torralba.
\newblock Multimodal neurons in pretrained text-only transformers.
\newblock In {\em {IEEE/CVF} International Conference on Computer Vision, {ICCV} 2023 - Workshops, Paris, France, October 2-6, 2023}, pages 2854--2859. {IEEE}, 2023.

\bibitem{sharma2018conceptual}
Piyush Sharma, Nan Ding, Sebastian Goodman, and Radu Soricut.
\newblock Conceptual captions: A cleaned, hypernymed, image alt-text dataset for automatic image captioning.
\newblock In Iryna Gurevych and Yusuke Miyao, editors, {\em Proceedings of the 56th Annual Meeting of the Association for Computational Linguistics (Volume 1: Long Papers)}, pages 2556--2565, Melbourne, Australia, 2018. Association for Computational Linguistics.

\bibitem{shenenhancing}
Shufan Shen, Zhaobo Qi, Junshu Sun, Qingming Huang, Qi~Tian, and Shuhui Wang.
\newblock Enhancing pre-trained representation classifiability can boost its interpretability.
\newblock In {\em The Thirteenth International Conference on Learning Representations, {ICLR} 2025, Singapore, April 24-28, 2025}. OpenReview.net, 2025.

\bibitem{shen2024expanding}
Shufan Shen, Junshu Sun, Xiangyang Ji, Qingming Huang, and Shuhui Wang.
\newblock Expanding sparse tuning for low memory usage.
\newblock In Amir Globersons, Lester Mackey, Danielle Belgrave, Angela Fan, Ulrich Paquet, Jakub~M. Tomczak, and Cheng Zhang, editors, {\em Advances in Neural Information Processing Systems 38: Annual Conference on Neural Information Processing Systems 2024, NeurIPS 2024, Vancouver, BC, Canada, December 10 - 15, 2024}, 2024.

\bibitem{shu2025large}
Dong Shu, Haiyan Zhao, Jingyu Hu, Weiru Liu, Ali Payani, Lu~Cheng, and Mengnan Du.
\newblock Large vision-language model alignment and misalignment: A survey through the lens of explainability.
\newblock {\em ArXiv preprint}, abs/2501.01346, 2025.

\bibitem{shukor2024implicit}
Mustafa Shukor and Matthieu Cord.
\newblock Implicit multimodal alignment: On the generalization of frozen llms to multimodal inputs.
\newblock {\em ArXiv preprint}, abs/2405.16700, 2024.

\bibitem{socher2013recursive}
Richard Socher, Alex Perelygin, Jean Wu, Jason Chuang, Christopher~D. Manning, Andrew Ng, and Christopher Potts.
\newblock Recursive deep models for semantic compositionality over a sentiment treebank.
\newblock In David Yarowsky, Timothy Baldwin, Anna Korhonen, Karen Livescu, and Steven Bethard, editors, {\em Proceedings of the 2013 Conference on Empirical Methods in Natural Language Processing}, pages 1631--1642, Seattle, Washington, USA, 2013. Association for Computational Linguistics.

\bibitem{pmlr-v202-sun23k}
Junshu Sun, Shuhui Wang, Xinzhe Han, Zhe Xue, and Qingming Huang.
\newblock All in a row: Compressed convolution networks for graphs.
\newblock In {\em Proceedings of the 40th International Conference on Machine Learning}, volume 202 of {\em Proceedings of Machine Learning Research}, pages 33061--33076. PMLR, 23--29 Jul 2023.

\bibitem{NEURIPS2024_93b7e278}
Junshu Sun, Chenxue Yang, Xiangyang Ji, Qingming Huang, and Shuhui Wang.
\newblock Towards dynamic message passing on graphs.
\newblock In {\em Advances in Neural Information Processing Systems}, volume~37, pages 80936--80964. Curran Associates, Inc., 2024.

\bibitem{team2023gemini}
Gemini Team, Rohan Anil, Sebastian Borgeaud, Jean-Baptiste Alayrac, Jiahui Yu, Radu Soricut, Johan Schalkwyk, Andrew~M Dai, Anja Hauth, Katie Millican, et~al.
\newblock Gemini: a family of highly capable multimodal models.
\newblock {\em ArXiv preprint}, abs/2312.11805, 2023.

\bibitem{touvron2023llama}
Hugo Touvron, Louis Martin, Kevin Stone, Peter Albert, Amjad Almahairi, Yasmine Babaei, Nikolay Bashlykov, Soumya Batra, Prajjwal Bhargava, Shruti Bhosale, et~al.
\newblock Llama 2: Open foundation and fine-tuned chat models.
\newblock {\em ArXiv preprint}, abs/2307.09288, 2023.

\bibitem{tu2023visual}
Cheng{-}Hao Tu, Zheda Mai, and Wei{-}Lun Chao.
\newblock Visual query tuning: Towards effective usage of intermediate representations for parameter and memory efficient transfer learning.
\newblock In {\em {IEEE/CVF} Conference on Computer Vision and Pattern Recognition, {CVPR} 2023, Vancouver, BC, Canada, June 17-24, 2023}, pages 7725--7735. {IEEE}, 2023.

\bibitem{van2008visualizing}
Laurens Van~der Maaten and Geoffrey Hinton.
\newblock Visualizing data using t-sne.
\newblock {\em Journal of machine learning research}, 9(11), 2008.

\bibitem{wah2011caltech}
Catherine Wah, Steve Branson, Peter Welinder, Pietro Perona, and Serge Belongie.
\newblock The caltech-ucsd birds-200-2011 dataset.
\newblock 2011.

\bibitem{wang2024qwen2}
Peng Wang, Shuai Bai, Sinan Tan, Shijie Wang, Zhihao Fan, Jinze Bai, Keqin Chen, Xuejing Liu, Jialin Wang, Wenbin Ge, et~al.
\newblock Qwen2-vl: Enhancing vision-language model's perception of the world at any resolution.
\newblock {\em ArXiv preprint}, abs/2409.12191, 2024.

\bibitem{xiao2010sun}
Jianxiong Xiao, James Hays, Krista~A. Ehinger, Aude Oliva, and Antonio Torralba.
\newblock {SUN} database: Large-scale scene recognition from abbey to zoo.
\newblock In {\em The Twenty-Third {IEEE} Conference on Computer Vision and Pattern Recognition, {CVPR} 2010, San Francisco, CA, USA, 13-18 June 2010}, pages 3485--3492. {IEEE} Computer Society, 2010.

\bibitem{xu2023energy}
Xinyue Xu, Yi~Qin, Lu~Mi, Hao Wang, and Xiaomeng Li.
\newblock Energy-based concept bottleneck models: Unifying prediction, concept intervention, and probabilistic interpretations.
\newblock In {\em The Twelfth International Conference on Learning Representations, {ICLR} 2024, Vienna, Austria, May 7-11, 2024}. OpenReview.net, 2024.

\bibitem{yuksekgonul2022post}
Mert Y{\"{u}}ksekg{\"{o}}n{\"{u}}l, Maggie Wang, and James Zou.
\newblock Post-hoc concept bottleneck models.
\newblock In {\em The Eleventh International Conference on Learning Representations, {ICLR} 2023, Kigali, Rwanda, May 1-5, 2023}. OpenReview.net, 2023.

\bibitem{espinosa2022concept}
Mateo~Espinosa Zarlenga, Pietro Barbiero, Gabriele Ciravegna, Giuseppe Marra, Francesco Giannini, Michelangelo Diligenti, Zohreh Shams, Fr{\'{e}}d{\'{e}}ric Precioso, Stefano Melacci, Adrian Weller, Pietro Li{\'{o}}, and Mateja Jamnik.
\newblock Concept embedding models: Beyond the accuracy-explainability trade-off.
\newblock In Sanmi Koyejo, S.~Mohamed, A.~Agarwal, Danielle Belgrave, K.~Cho, and A.~Oh, editors, {\em Advances in Neural Information Processing Systems 35: Annual Conference on Neural Information Processing Systems 2022, NeurIPS 2022, New Orleans, LA, USA, November 28 - December 9, 2022}, 2022.

\bibitem{zhang2025flexvln}
Siqi Zhang, Yanyuan Qiao, Qunbo Wang, Longteng Guo, Zhihua Wei, and Jing Liu.
\newblock Flexvln: Flexible adaptation for diverse vision-and-language navigation tasks.
\newblock {\em arXiv preprint arXiv:2503.13966}, 2025.

\bibitem{zhang2025cosmo}
Siqi Zhang, Yanyuan Qiao, Qunbo Wang, Zike Yan, Qi~Wu, Zhihua Wei, and Jing Liu.
\newblock Cosmo: Combination of selective memorization for low-cost vision-and-language navigation.
\newblock {\em arXiv preprint arXiv:2503.24065}, 2025.

\bibitem{zhao2024gradient}
Chenyang Zhao, Kun Wang, Xingyu Zeng, Rui Zhao, and Antoni~B Chan.
\newblock Gradient-based visual explanation for transformer-based clip.
\newblock In {\em International Conference on Machine Learning}, pages 61072--61091. PMLR, 2024.

\end{thebibliography}
